\documentclass[letterpaper,12pt]{article}

\usepackage{amsmath} 
\usepackage[american]{babel}
\usepackage{tabularx}
\usepackage{booktabs}
\pdfoutput=1 
\usepackage{graphicx}
\usepackage{array}
\usepackage{tabu}
\usepackage{mathrsfs}

\newcommand\T{\rule{0pt}{2.6ex}}       
\newcommand\B{\rule[-1.2ex]{0pt}{0pt}} 

\usepackage{xfrac}

\usepackage{framed, color}
\usepackage{mdframed, color}

    \usepackage[utf8]{inputenc}
    \usepackage[T1]{fontenc}
    \usepackage{lmodern}
    \usepackage[x11names]{xcolor}
    \usepackage{framed}
    \colorlet{shadecolor-pink}{LavenderBlush2}
    \colorlet{framecolor}{Red1}
    \usepackage{lipsum}

    {\endMakeFramed}

    \newenvironment{frshaded*}{%
    \MakeFramed {\advance\hsize-\width \FrameRestore}}%
    {\endMakeFramed}

\usepackage{tcolorbox}

\usepackage{amsthm,color}
\theoremstyle{definition}

\mdfdefinestyle{example1}{%
    linecolor=blue,
    outerlinewidth=2pt,
    bottomline=false,
    leftline=false,rightline=false,
    skipabove=\baselineskip,
    skipbelow=\baselineskip,
    frametitle=\mbox{},
}

\mdfdefinestyle{exampledefault}{%
rightline=true,
leftline = true,
innerleftmargin=10,innerrightmargin=10,
frametitlerule=true,frametitlerulecolor=green,
frametitlebackgroundcolor=yellow,
frametitlerulewidth=2pt}

\usepackage{url}            
\usepackage{booktabs}       
\usepackage{amsfonts}       
\usepackage{nicefrac}       
\usepackage{microtype}      

\definecolor {shadecolor}{rgb}{1, 0.8, 0.3}
\definecolor {eqn-descrpt-aqua}{RGB}{210, 240, 235}
\definecolor {eqn-descrpt-frame-aqua}{RGB}{83, 111, 121}
\definecolor {med-aqua}{RGB}{165, 210, 200}
\definecolor {eqn-celadon}{RGB}{215, 240, 221}
\definecolor {eqn-frame-celadon}{RGB}{33, 130, 125}

\title{A Variational Approach to Parameter Estimation for Characterizing 2-D Cluster Variation Method Topographies}
\vspace{10 mm}

\author{Alianna J. Maren$^{1,2,\dagger}$ \\
  \vspace{1 mm} \\ 
 $^1$ Themesis, Inc.  \\
  {\small 3-2600 Kaumualii Hwy Ste 1300 PMB 188, Lihue, HI 96766} \\
   \vspace{1 mm} \\
$^2$  Northwestern University School of Professional Studies\\
  Master of Science in Data Science Program\\ 
  {\small 633 Clark St, Evanston, IL 60208} \\  
  \vspace{10 mm} \\ 
  {$\dagger$  \small {Address to which correspondence should be addressed:}}\\
  {\tt themesisinc1@gmail.com} \\
  \vspace{1 mm} \\   
  {\tt alianna.maren@northwestern.edu}\\
  \vspace{10 mm} \\ 
  Themesis Technical Report THM  TR2022-001  (ajm)\\

\date{Revision Date: 2022-09-08\\
Version 1.0}
  }

\begin{document}

\maketitle
\thispagestyle{empty} 

\newpage

\setcounter{page}{1}

\abstract{One of the biggest challenges in characterizing 2-D topographies is succinctly communicating the dominant nature of local configurations. In a 2-D grid composed of bistate units, this could be expressed as finding the characteristic configuration variables such as nearest-neighbor pairs and triplet combinations. The 2-D cluster variation method (CVM)  provides a theoretical framework for associating a set of configuration variables with only two parameters, for a system that is at free energy equilibrium. This work presents a method for determining which of many possible two-parameter sets provides the ``most suitable'' match for a given 2-D topography, drawing from methods used for variational inference. 

This particular work focuses exclusively on topographies for which the activation enthalpy parameter ($\varepsilon_0$) is zero, so that the distribution between two states is equiprobable. This condition is used since, when the two states are equiprobable, there is an analytic solution giving the configuration variable values as functions of the \textit{h-value}, where we define \textit{h} in terms of the interaction enthalpy parameter ($\varepsilon_1$) as $h = exp(2\varepsilon_1)$. This allows the computationally-achieved configuration variable values to be compared with the analytically-predicted values for a given \textit{h-value}. 

The method is illustrated using four patterns derived from three different naturally-occurring black-and-white topographies, where each pattern meets the equiprobability criterion. 

We achieve expected results, that is, as the patterns progress from having relatively low numbers of like-near-like nodes to increasing like-near-like masses, the \textit{h-values} for each corresponding free energy-minimized model also increase. Further, the corresponding configuration variable values for the (free energy-minimized) model patterns are in close alignment with the analytically-predicted values. 

The method described here has applicability beyond characterizing specific 2-D topographies. Potential applications extend to active inference as well as to 2-D CORTECONs, which incorporate free energy minimization into a 2-D grid of latent variables, in addition to the usual methods employed with energy-based neural networks.}

\vspace{10pt}


\pagebreak

%
\section{Introduction and Overview}
\label{sec:intro-and-overview}
%

Finding a free energy-based approach to modeling a two-dimensional (2-D) topography in terms of local \textit{configuration variables} (nearest-neighbor, next-nearest-neighbor, and triplets) is an interesting challenge because the model itself must be brought to a free energy minimum for a given set of enthalpy parameters. Once this has been done, for each set of enthalpy parameters, it would be possible to measure the divergence between the model's at-equlibrium configuration variable values as compared with those in the actual topography being modeled. 

\begin{figure}[ht]
  \centering
  \fbox{
  \rule[-.5cm]{0cm}{4cm}\rule[-.5cm]{0cm}{0cm}	
  \includegraphics [trim=0.0cm 0cm 0.0cm 0cm, clip=true,   width=0.95\linewidth]{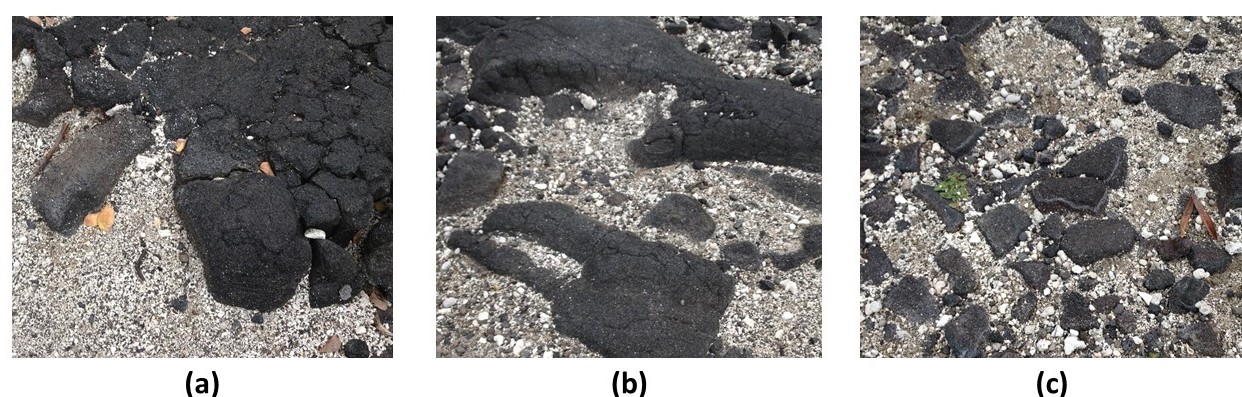}} 
  \vspace{3mm} 
  \caption{Illustration of three naturally-occurring topographies, observed within several yards of each other on the 1871 Trail in the Pu{`}uhonua o H\={o}naunau National Historical Park near Captain Cook, on~the Big Island of Hawai{`}i: {(\textbf{a})} Larger lava rocks surrounded by white coral sand, where all lava rock elements are connected with each other. {(\textbf{b})} A more fragmented version, in~which various distinct sizes of lava rocks appear, yielding a range of lava rock sizes and varied degrees of connectiveness. {(\textbf{c})} A yet more granular version of the same, with~coral fragments approaching the size of some of the lava pebbles. Photos by A.J. Maren, 2020.}   
\label{fig:Rocks-three-topographies-horiz-crppd-2020-11-16}
\end{figure}

This conundrum - that of finding the set of enthalpy parameters yielding the ``best-fit'' to a specific topography - is illustrated in Figure~\ref{fig:Rocks-three-topographies-horiz-crppd-2020-11-16}, which shows a series of three naturally-occurring topographies. Each of the three images shown in Figure~\ref{fig:Rocks-three-topographies-horiz-crppd-2020-11-16} presents a specific terrain, where all three terrains are located within several dozen meters of each other, and the images were taken with the same camera on the same day, within minutes of each other. 

\vspace{3mm} 

The three naturally-occurring topographies shown in Figure~\ref{fig:Rocks-three-topographies-horiz-crppd-2020-11-16} offer different granularities of solid masses of black lava rocks surrounded by white coral sand, where the lava rock elements range from most contiguous in  Figure~\ref{fig:Rocks-three-topographies-horiz-crppd-2020-11-16}\textbf{(a)} to least contiguous (most granular) in  Figure~\ref{fig:Rocks-three-topographies-horiz-crppd-2020-11-16}\textbf{(c)}.

The challenge lies not so much with establishing a 2-D modeling system that can be brought to a free energy equilibrium. The 2-D cluster variation method (2-D CVM), originally proposed by Kikuchi (1951) \cite{Kikuchi_1951_Theory-coop-phenomena} and further evolved by Kikuchi and Brush (1967) \cite{Kikuchi-Brush_1967_Improv-CVM}, is appropriate for this task. 

Instead, the real unsolved problem lies in determining which parameter set, taken from a set of  2-D CVM models, each constructed using a specific set of enthalpy parameters, provides a ``best fit''  for a given initial topography. 

This Technical Report presents a new methodology addressing this question, introducing a new divergence method that has its conceptual origins in the Kullback-Leilber divergence, yet is specialized to address topographies that can be modeled using the 2-D CVM. It presents results for the example set of images shown in Figure~\ref{fig:Rocks-three-topographies-horiz-crppd-2020-11-16}, where the  parameter values follow the expected progression. 

This Report further serves three purposes:

\begin{itemize}
\setlength{\itemsep}{1pt}
\item \textbf{\textit{Extended and self-contained}}, including material that will not be included in the journal paper that will be based on this work, providing a single and reasonably comprehensive source for anyone wishing to understand and replicate the results, 
\item \textbf{\textit{Includes essential background materials}}, reducing the need to consult prior works, and 
\item \textbf{\textit{Substantial details on the technical results}}, including details of how sets of model-based configuration variables (describing local pattern characteristics) compare with the configuration variable values for each of the original, natural images. This is done across a range of different enthalpy parameters, making it possible to identify distinct sets of enthalpy parameters, each yielding the ``best fit'' for a specific natural terrain. 
\end{itemize}

The remaining sections describe how the three topographies shown in Figure~\ref{fig:Rocks-three-topographies-horiz-crppd-2020-11-16} can be modeled using the 2-D CVM.  

The following Table~\ref{tbl:glossary-thermodynamic} presents a glossary of the thermodynamic terms used in this Report.

%
\begin{table}[ht!]\footnotesize
    \caption{Thermodynamic Variable Definitions}  
    \label{tbl:glossary-thermodynamic}
    \centering 
    \vspace{3mm}
    \begin{tabular}{|p{3cm}|p{10cm}|}
    \hline
	 \multicolumn{1}{|>{\centering\arraybackslash}m{3cm}|}	{\textbf{Variable}} 
    & \multicolumn{1}{>{\centering\arraybackslash}m{10cm}|}{\textbf{Meaning}} \T\B \\ 
    \hline                        
	 \multicolumn{1}{|>{\centering\arraybackslash}m{3cm}|}	{Activation enthalpy} 
    &Enthalpy  $\varepsilon_0$ associated with a single unit (node) in the ``on'' or ``active'' state (\textbf{A}); influences configuration variables and is set to 0 (for this work) in order to achieve an analytic solution for the free energy equilibrium  \\ [3pt] 		

	 \multicolumn{1}{|>{\centering\arraybackslash}m{3cm}|}	{Configuration variable(s)} 
    &Nearest neighbor, next-nearest neighbor, and triplet patterns  \\ [8pt] 

	 \multicolumn{1}{|>{\centering\arraybackslash}m{3cm}|}	{Degeneracy} 
    &Number of ways in which a configuration variable can appear  \\ [5pt] 

	 \multicolumn{1}{|>{\centering\arraybackslash}m{3cm}|}	{Enthalpy} 
    &Internal energy \textit{H} results from both per unit and pairwise interactions; often denoted $H$ in thermodynamic treatments    \\ [5pt]  
    
	 \multicolumn{1}{|>{\centering\arraybackslash}m{3cm}|}	{Entropy} 
    &The entropy \textit{S} is the distribution over all possible states; often denoted $S$ in thermodynamic treatments and $H$ in information theory \\ [5pt]       

	 \multicolumn{1}{|>{\centering\arraybackslash}m{3cm}|}	{Equilibrium point} 
    &By definition, the free energy minimum for a closed system  \\ [5pt] 
 
	 \multicolumn{1}{|>{\centering\arraybackslash}m{3cm}|}	{Equilibrium distribution} 
    & Configuration variable values when free energy minimized for given \textit{h-value}  \\ [5pt]   

	 \multicolumn{1}{|>{\centering\arraybackslash}m{3cm}|}	{Ergodic distribution} 
    & Achieved when a system is allowed to evolve over a long period of time   \\ [5pt]  
 
	 \multicolumn{1}{|>{\centering\arraybackslash}m{3cm}|}	{Free Energy} 
    &The thermodynamic state function \textit{F}; where \textit{F = H-TS}; sometimes \textit{G} is used instead of \textit{F}; referring to (thermodynamic) Gibbs free energy   \\ [5pt]       
    
	 \multicolumn{1}{|>{\centering\arraybackslash}m{3cm}|}	{\textit{h-value}} 
    & A more useful expression for the interaction enthalpy parameter $\varepsilon_1$; $h = e^{2\beta\varepsilon_1}$, where $\beta = 1/{k_{\beta}T}$,  and where $k_{\beta}$ is Boltzmann's constant and $T$ is temperature; $\beta$ can be set to 1 for our purposes   \\ [5pt]       

	 \multicolumn{1}{|>{\centering\arraybackslash}m{3cm}|}	{Interaction enthalpy} 
    & Between two unlike units, $\varepsilon_1$; influences configuration variables  \\ [7pt] 
 
	 \multicolumn{1}{|>{\centering\arraybackslash}m{3cm}|}	{Interaction enthalpy parameter} 
    & Another term for the \textit{h-value} where $h=e^{2\varepsilon_1}$  \\ [9pt]   
    
	 \multicolumn{1}{|>{\centering\arraybackslash}m{3cm}|}	{Temperature} 
    &Temperature \textit{T} times Boltzmann's constant $k_{\beta}$ is set equal to one \\ [10pt]    
     \hline	  
  \end{tabular}
\end{table}
%

The remainder of this paper is organized into the following sections:

\begin{itemize}
\item Section 2 identifies the goals and expectations for this work, with a focus on the expected behavior and values-range for the crucial interaction enthalpy parameter \textit{h-value}, and further places the parameter study in the context of variational Bayes.
\item Section 3 provides a detailed walkthrough of a single prior result, to establish context and perspective for this work. It further identifies the \textit{interpretation variables} that are used throughout this, as well as prior, work; showing how system behavior can be characterized just by examining three variables.  
\item Sections 4 and 5 provide background on variational Bayes (together with active inference) and the 2-D CVM, respectively. 
\item Section~\ref{sec:config-variables} briefly overviews the configuration variables used by Kikuchi and Brush, and in this work.
\item Section~\ref{sec:2-D-CVM} reviews the 2-D CVM method itself, presenting the essential free energy equation, together with the analytic solution available when there is an equiprobable distribution of nodes into states \textbf{A} and \textbf{B}. 
\item Section 8 presents the data used, Section 9 overviews the methods, and Section 10 presents the results. 
\item Section 11 discusses the results in terms of potential applications, and Section 12 provides a summary. 
\item Section 13 identifies future directions. 
\end{itemize}

%
\section{Research Goals and Expectations}
\label{sec:rsch-goals-and-expects}
%

The CVM approach is a useful and insightful way for describing systems in which the entropy is considered to be more than the relative proportion of ``on'' and ``off'' units. (Throughout this work, and in both prior and subsequent works, the term ``nodes'' will be used interchangeably with ``units.'') In short, the CVM approach allows us to address the entropy of \textit{patterns}, and not just the simple entropy associated with the fractions of units ($x_1$ and $x_2$) in each of the two energy states of a bistate system. The nature of these local patterns is captured via a set of \textit{configuration variables}, which will be fully discussed later in this Report.

Only two parameters are needed to specify a 2-D CVM system; the activation enthalpy $\varepsilon_0$ and the interaction enthalpy $\varepsilon_1$. 

This work  is confined to the case where the activation enthalpy $\varepsilon_0 = 0$, thus ensuring an equiprobable distribution of nodes into states \textbf{A} and  \textbf{B} ($x_1 = x_2 = 0.5$). When this condition holds, there is an analytic solution for the configuration variable values in terms of $\varepsilon_1$.

Due to the form of this analytic solution, we find it convenient to use a parameter termed the \textit{h-value} (or simply, $h$), where the \textit{h-value} is a function of the  interaction enthalpy parameter $\varepsilon_1$. Specifically, $h = e^{2\beta\varepsilon_1}$, where $\beta = 1/{k_{\beta}T}$,  and where $k_{\beta}$ is Boltzmann's constant and $T$ is temperature; we can set $\beta = 1$ for our purposes.

%
\subsection{Specific Parameter Expectations}
\label{subsec:specific-param-expects}
%

This work presents a method for finding the \textit{h-value} associated with a model that yields the ``most suitable'' or ``best fit'' correspondence to the representation of a given 2-D topography. We advance a new divergence measure, specific to work with the 2-D CVM, that allows determination of the enthalpy parameter(s) providing this ``best fit.''

Even before we advance a specific method as to how a given ``most suitable'' \textit{h-value} would be found, we can form some initial expectations based on visual examination of the three topographies presented in Figure~\ref{fig:Rocks-three-topographies-horiz-crppd-2020-11-16}. 

As we progress left-to-right, from Figure~\ref{fig:Rocks-three-topographies-horiz-crppd-2020-11-16}\textbf{(a)} to Figure~\ref{fig:Rocks-three-topographies-horiz-crppd-2020-11-16}\textbf{(c)}, we see that there are  more contiguous areas on the left, and the terrain becomes more granular as we move to the right. 

The interaction enthalpy parameter $\varepsilon_1$ (and correspondingly, the \textit{h-value}) governs the extent to which like-near-like units gravitate towards each other in a free energy-minimized system. That is, when $\varepsilon_1 = 0$ (or $h=1$), there is no ``interaction enthalpy'' between units; the free energy is not further minimized by bringing ``like'' nodes together. The resultant system takes on a very granular form, where the appearance of of any ``masses'' of units in either state \textbf{A} or state \textbf{B} is a matter of chance, much as it would be random chance to get a long string of either ``heads'' or ``tails'' in a series of random coin flips. 

This situation, of having a very low \textit{h-value} ($h \approx 1$) would be illustrated by the terrain in Figure~\ref{fig:Rocks-three-topographies-horiz-crppd-2020-11-16}\textbf{(c)}.

On the other hand, when we have $\varepsilon_1 > 1$, we create a propensity towards like-near-like aggregation. This is shown, progressively, in Figure~\ref{fig:Rocks-three-topographies-horiz-crppd-2020-11-16}\textbf{(b)}, and more so in Figure~\ref{fig:Rocks-three-topographies-horiz-crppd-2020-11-16}\textbf{(a)}, where the black lava rocks are massed together. 

Thus, we would expect that our method for providing us with enthalpy parameters yielding ``best fit'' 2-D CVM models would give us very low \textit{h-values} for systems such as those shown in Figure~\ref{fig:Rocks-three-topographies-horiz-crppd-2020-11-16}\textbf{(c)} ($h \approx 1$), and higher \textit{h-values} ($h >> 1$) for systems such as those shown in Figure~\ref{fig:Rocks-three-topographies-horiz-crppd-2020-11-16}\textbf{(a)}. 

Further, we would expect that the evolution of \textit{h-values} would be smooth and continuous, as the configuration variable values describing different terrains change smoothly across terrains.

%
\subsection{Role of Variational Bayes}
\label{subsec:role-var-Bayes}
%

As beautifully expressed by Beal (2014) \cite{Beal_2003_Variational-algorithm-approx-Bayes-inference}, ``The goal of variational inference is to approximate a conditional density of latent variables given observed variables.'' 

This work adopts a variational approach to finding the parameters yielding the ``best fit'' or ``best approximation'' to a data representation expressed using a 2-D grid of offset units, and continues work initially developed by Maren (2019a) \cite{ Maren_2019_Deriv-var-Bayes}.   

In the case of working with the 2-D CVM, the observed variables are the configuration variables, that is, the relative fractions of not only the units in states \textbf{A} and \textbf{B}, but also the nearest-neighbor and next-nearest-neighbor pairs, as well as the triplets. 

The latent variables for the 2-D CVM are the pair of enthalpy parameters; $(\varepsilon_0, \varepsilon_1)$, or since we prefer to work with the \textit{h-value},  $(\varepsilon_0, h)$.

For the specific study addressed in this work, we keep $\varepsilon_0 = 0$, so we are only interested in determining a given $(h)$, or more precisely (in Bayesian germs), $(h|C)$, where $C$ is the entire set of configuration variables, which will be discussed later in this work.

%
\subsection{Potential Uses and Applications}
\label{subsec:uses-and-applications}
%

The essential notion of the cluster variation method, or CVM, is that we work with a more complex entropy expression than is typically used for describing the free energy of a bistate system. 

Up until now, the CVM approach has not received a great deal of attention. However, as Figure~\ref{fig:2D-CVM-init-and-FEMin-scale-free_h-eq-1pt165_crppd_2019-06-26} illustrates, this approach is potentially useful for characterizing 2-D systems that would naturally gravitate to a free energy-minimized state. Examples of potential applications include: 

\begin{itemize}
\setlength{\itemsep}{1pt}
\item \textbf{\textit{Modeling systems where the topography is an essential component of system description}}; this can range from urban/rural topographies to medical images, 
\item \textbf{\textit{Providing an essential component of a new computational engine}}, where \textit{memories} of prior system states can persist over time, exemplified by the CORTECON neural network architectures (where CORTECON stands for \textit{COntent-Retentive TEmporally-CONnected}) introduced by Maren et al. in 1992 \cite{Maren_1992_Free-energy-as-driving-function, Maren-Schwartz-Seyfried_1992_Config-entropy-stabilizes, Schwartz-Maren_1994_Domains-interacting-neurons}, and continued in 2015 \cite{AJMaren-HSzu-New-EEG-Measure-SPIE-STA-2015}, and 
\item \textbf{\textit{Providing the modeling component of a variational Bayes approach}} that uses a \textit{representational} system to model an \textit{external} system, where the two are separated by a Markov blanket, and both systems are presumed to come to separate free energy equilibrium states; we particularly envision a role for the 2-D CVM in active inference (see, e.g., Sajid et al. (2020) \cite{Sajid-and-Friston-et-al_2020_Active-Inference}). 
\end{itemize}

Maren (2019b, 2021) has previously applied the 2-D CVM free energy minimization to two different 2-D topographies \cite{Maren_2019_2D-CVM-FE-fundamentals-and-pragmatics, Maren_2021_2-D-CVM-Topographies}. This work builds on that pair of conjoined works, making this essentially the second in a series addressing 2-D CVM topographies and correlating phase space parameters $(\varepsilon_0, h)$ with configuration variables expressing local patterns.

%
\section{Example of Prior Results}
\label{sec:example-prior-results}
%

It will help ground the work done in this study, and establish a perspective for what will be done here, to first examine previously-obtained results. To do this, we refer to an exemplar free energy-minimized 2-D CVM system, using  a manually-designed topography as a starting point (Maren, 2019b, 2021) \cite{Maren_2019_2D-CVM-FE-fundamentals-and-pragmatics, Maren_2021_2-D-CVM-Topographies}. These results,  shown in the following Figure~\ref{fig:2D-CVM-init-and-FEMin-scale-free_h-eq-1pt165_crppd_2019-06-26}, were initially presented as as Figure 1 of both Maren (2019b, 2021). 

Figure~\ref{fig:2D-CVM-init-and-FEMin-scale-free_h-eq-1pt165_crppd_2019-06-26} illustrates the case for which \textbf{(a)} a manually-designed initial system on the LHS (Left-Hand-Side) has been \textbf{(b)} brought to free energy equilibrium on the RHS (Right-Hand-Side). 

\begin{figure}[ht]
  \centering
  \fbox{
  \rule[-.5cm]{0cm}{4cm}\rule[-.5cm]{0cm}{0cm}	
  \includegraphics [trim=0.0cm 0cm 0.0cm 0cm, clip=true,   width=0.95\linewidth]{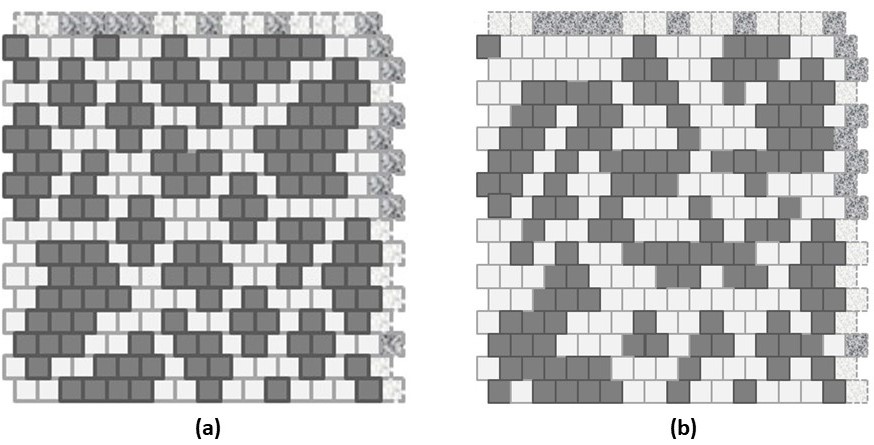}} 
  \vspace{3mm} 
  \caption{Illustration of \textbf{(a)} a manually-designed 2-D CVM grid that is \textbf{(b)} brought to a free energy equilibrium configuration for $h=1.65$, where $h = exp(2 \varepsilon_1)$.}   
\label{fig:2D-CVM-init-and-FEMin-scale-free_h-eq-1pt165_crppd_2019-06-26}
\end{figure}
\vspace{3mm} 

The at-equilibrium results presented in the RHS of Figure~\ref{fig:2D-CVM-init-and-FEMin-scale-free_h-eq-1pt165_crppd_2019-06-26} show a combination of topographic elements, which we describe as ``spiderlegs'' together with both ``rivers'' and ``peninsulas.'' An example of each of these is on the upper-left-hand-side of  Figure~\ref{fig:2D-CVM-init-and-FEMin-scale-free_h-eq-1pt165_crppd_2019-06-26}~\textbf{(b)}. 

In this figure, several sets of ``on'' (\textbf{A}, or black) nodes that initially were ``islands'' of from four to eight nodes have been morphed to form a single, more extensive ``landmass.'' This resultant ``landmass,'' which connects several previously-isolated ``islands,'' is characterized by extended diagonal elements that are either one or two nodes in width (the ``spiderlegs''). When the black nodes (\textbf{A}) form these extended legs against the background of white (\textbf{B}) nodes, we see elements that visually appear to be ``peninsulas.'' Interestingly, the peninsula of \textbf{B} nodes in the upper left-hand-corner of Figure~\ref{fig:2D-CVM-init-and-FEMin-scale-free_h-eq-1pt165_crppd_2019-06-26}~\textbf{(b)} is penetrated, through its full length (saving a single connection node at the top), by a long ``river'' of white or \textbf{A} nodes. 

The prior work, yielding the results shown in Figure~\ref{fig:2D-CVM-init-and-FEMin-scale-free_h-eq-1pt165_crppd_2019-06-26}, was limited in that it used manually-designed patterns as its origination topographies. These manually-designed patterns were each brought to a free energy equilibrium, using a specific \textit{h-value} in each case. However, these initial patterns were far from an equilibrium configuration, so that the \textit{h-values} that were used were visual estimates of what would yield a free energy-minimized system resultant from the initiating pattern. 

Thus, there were three challenges that confronted the prior work:

\begin{itemize}
\item The manually-designed initiating patterns were far from equilibrium, so that the configuration varaibles for those patterns did not yield a single corresponding target \textit{h-value}, 
\item Because the initiating patterns were so far from equilibrium, it was not possible (using the simple node-flipping algorithm for free energy minimization) to bring the system to true equilibrium within a reasonable timeframe, so the resultant system - although closer to an equilibrium state (as measured by a decrease in free energy), was still not at a free energy equilibrium,
\item It was not possible (given the methods in the earlier study) to determine which \textit{h-value} gave the ``best'' results, in terms of bringing the system to an equilibrium state.  
\end{itemize}

This work rectifies those problems, by:

\begin{itemize}
\item Using initiating patterns drawn from nature, with the hope and anticipation that they will each be closer to equilibrium and can be readily brought to an equilibrium state, and
\item Introducing a method - a new divergence measure - that provides a means for determining which \textit{h-value} yields the free energy minimized state that is \textit{closest} to the initiating pattern. 
\end{itemize}

Despite the challenges in the previous work (Maren (2021) \cite{Maren_2021_2-D-CVM-Topographies}), some very interesting topographic features emerged, e.g. the ``spider-legs,'' ``channels,'' and ``rivers' that were previously mentioned. One goal of this work and anticipated future studies will be to further investigate the appearance of these topographic features and to identifiy their dependence on the key parameters $(\varepsilon_0, h)$.

%
\subsection{The Interpretation Variables}
\label{subsec:interp-variables}
%

In order to get a simple and easy-to-visualze understanding of 2-D CVM topographies, we extract three variables from the total set of fourteen configuration variables. We refer to these as the \textit {interpretation variables}.

The fourteen configuration variables, which will be described fully in Section~\ref{sec:config-variables}, comprise the following: 

\begin{itemize}
\item Two single node fraction variables; $x_1$ and $x_2$, 
\item Three nearest-neighbor pair variables; the collective set of $y_i$, 
\item Three next-nearest-neighbor pairs variables; the collective set of $w_i$, and 
\item Six triplet variables; the collective set of $z_i$.  
\end{itemize}

We nominate three configuration variables to form our set of \textit{interpretation variables}: $y_2$, $z_3$, and~$z_1$:

\begin{itemize}
\item $y_2$ - the \textbf{A}-\textbf{B} nearest-neighbor pairs; indicates the relative extent to which the \textbf{A} units are distributed among the surrounding \textbf{B} units; a~higher $y_2$ value indicates lots of boundary areas between \textbf{A} and \textbf{B}, and~a smaller value indicates more compact ``landmasses`` of \textbf{A} units,
\item $z_3$ - the \textbf{A}-\textbf{B}-\textbf{A} triplets; indicates the relative fraction of \textbf{A} units that are involved in a ``jagged'' border (one that involves irregular protrusions of \textbf{A} into a \textbf{B} space), or~the presence of one or more thin ``rivers'' of \textbf{B} units extending into landmass(es) of \textbf{A} units, and
\item $z_1$ - the \textbf{A}-\textbf{A}-\textbf{A} triplets; indicates the relative fraction of \textbf{A} units that are included within the \textit{interiors} of the ``islands'' or ``land masses''; $z_1$ also (indirectly) indicates the compactness of these masses.

\end{itemize}

We understand, of~course, that  a given set of values for the \textit{interpretation variables} will not define a specific topography; rather, they will correlate with a specific \textit{kind of} topology.

%
\subsection{The Analytic Solution for the 2-D CVM-example}
\label{subsec:analytic-solution-example}
%

One important reason for working with the equiprobable distribution case, where $x_1 = x_2 = 0.5$ (and correspondingly, $\varepsilon_0 = 0$), is that for this case, we can find an analytic solution for the various configuration variables in terms of the single interaction enthalpy $\varepsilon_1$, or (more usefully), in terms of $h = exp(2\varepsilon_1)$. 

\begin{figure}[ht]
  \centering
  \fbox{
  \rule[-.5cm]{0cm}{4cm}\rule[-.5cm]{0cm}{0cm}	
  \includegraphics [trim=0.0cm 0cm 0.0cm 0cm, clip=true,   width=0.95\linewidth]{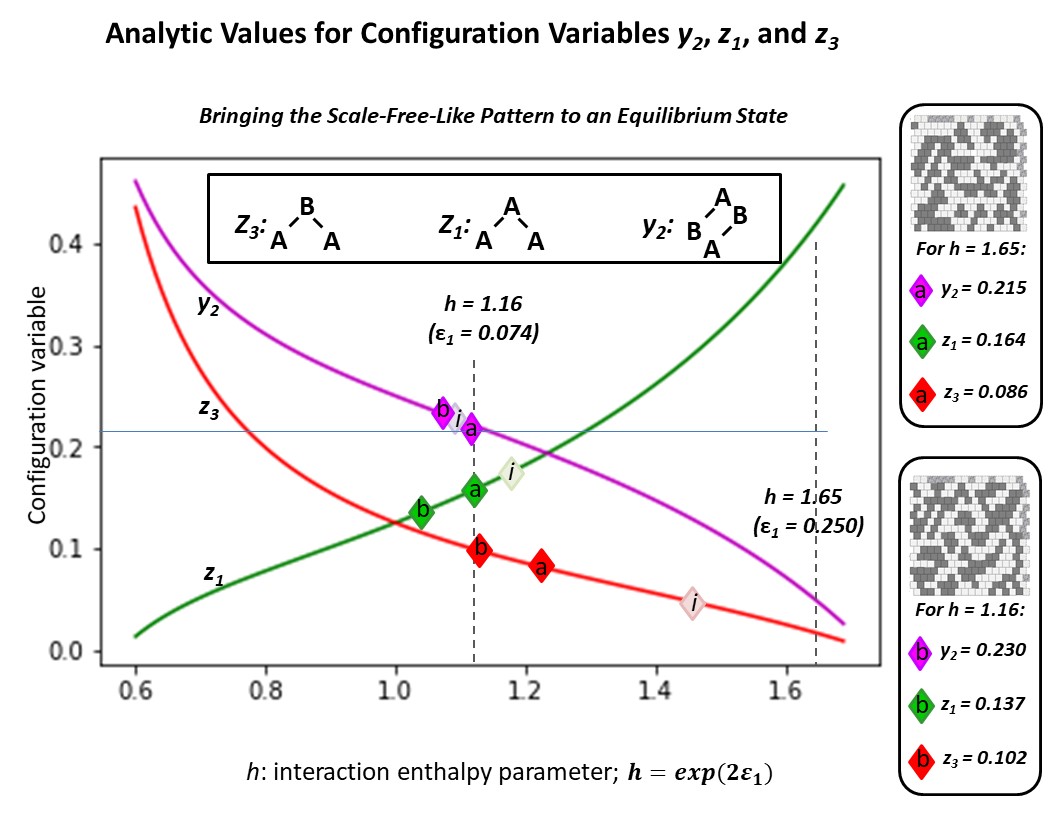}} 
  \vspace{3mm} 
  \caption{Illustration of computed configuration variable valuess when the initial pattern (shown in Figure~\ref{fig:2D-CVM-init-and-FEMin-scale-free_h-eq-1pt165_crppd_2019-06-26}  \textbf{(a)}) has been brought to equilibrium for each of two different \textit{h-values}; $h = 1.16$ and $h = 1.65$. See text for details.}   
\label{fig:2D-CVM_Expts-scale-free-h-eq-1pt65-rev_2021-02-16_scale-free-rslts}
\end{figure}

Figure~\ref{fig:2D-CVM_Expts-scale-free-h-eq-1pt65-rev_2021-02-16_scale-free-rslts} reproduces Figure~12 in Maren 2021 \cite{Maren_2021_2-D-CVM-Topographies}. It presents the graphs of $y_2$, $z_3$, and $z_1$ as functions of $h = exp(2\varepsilon_1$). It further shows the computationally-determined values for those interpretation variables, placed as dark diamonds on the corresponding function graphs, together with the initial values for those interpretation variables, all for the example presented earlier in Figure~\ref{fig:2D-CVM-init-and-FEMin-scale-free_h-eq-1pt165_crppd_2019-06-26}. The computationally-obtained interpretation values are obtained for two different \textit{h-values}; one where $h = 1.16$ and one where $h=1.65$.

\vspace{3mm} 

%
\subsubsection{The Interpretation Variables as Functions of \textit{h}}
\label{subsubsec:interp-var-functions}
%

To present a visual context for the results shown here in Figure~\ref{fig:2D-CVM_Expts-scale-free-h-eq-1pt65-rev_2021-02-16_scale-free-rslts}, we briefly examine the three graphs of the functions $y_2(h)$, $z_3(h)$, and $z_1(h)$. 

When $h=1$, then $\varepsilon_1 = 0$, as $h = exp(2\varepsilon_1)$. At this value, we see that the analytic graph shows that $y_2$ = 0.25 and $z_1 = z_3 = 0.125$. These are the values that would be expected when the distribution of nearest-neighbor pairs (the $y_i$) and the various triplets (the $z_i$) are random, as there is no interaction enthalpy to either pull like nodes together or to push them apart. Note that $y_2$ has a degeneracy factor of $2$ associated with it, as it can appear either as a \textbf{A}-\textbf{B} pair, or as a  \textbf{B}-\textbf{A} pair, read left-to-right. The value of $y_2 = 0.25$ is before multiplication by the degeneracy factor of $2$. Correspondingly, $y_1$ and $y_3$ (neither shown in this figure) also have values of $0.25$ when $h = 1.0$. (Details about the degeneracy factors, etc., are presented in Section~\ref{sec:config-variables}; \textit{Configuration Variables}). 

When $h<1$, then $\varepsilon_1 <0$. A negative interaction enthalpy means that like-near-unlike pairings (as evidenced by $y_2$ and $z_3$) lower the free energy. (See the free energy equation presented later in this work as Eqn.~\ref{eqn:Bar-H-2-D-basic-eqn}.) Thus, when we have $h<1$, then we also have $y_2 >0.25$ and $z_3 > 0.125$, and correspondingly, the fraction of same-type triplets ($z_1$) is decreased, or $z_1 < 0.125$. 

This is readily observed in Figure~\ref{fig:2D-CVM_Expts-scale-free-h-eq-1pt65-rev_2021-02-16_scale-free-rslts}, where we see (for example) that $z_1 \rightarrow 0$ as $h \rightarrow 0$ on the left-hand side of the figure. We interpret this by realizing that when $h < 1$, we would \textit{increase} the free energy by having like-near-like nodes. Thus, to minimize the free energy, we would push similar nodes apart from each other, reducing the $z_1$ value, corresponding to \textbf{A}-\textbf{A}-\textbf{A} triplets. 

Correspondingly, for the values of $h$ most of interest to us ($h>1.0$), we get increasing like-near-like triplets $z_1$, or $z_1 > 0.125$ when $h >1.0$. 

This means, that if we have a very agglomerative topography, we would expect a relatively high \textit{h-value}. If we have a topography in which there is some local cohesion, but which is still largely granular, then we expect that $h>1.0$, but not by as much.

%
\subsubsection{Computationally-Obtained Interpretation Variables}
\label{subsubsec:icomput-obtained-nterp-vars}
%

Figure~\ref{fig:2D-CVM_Expts-scale-free-h-eq-1pt65-rev_2021-02-16_scale-free-rslts} shows the results for two free energy minimization studies on the same initiating topography, presented earlier as Figure~\ref{fig:2D-CVM-init-and-FEMin-scale-free_h-eq-1pt165_crppd_2019-06-26} \textbf{(a)}.  

The interpretation variables corresponding to the original pattern (presented as Figure~\ref{fig:2D-CVM-init-and-FEMin-scale-free_h-eq-1pt165_crppd_2019-06-26} \textbf{(a)}) are shown in light-colored diamonds, placed on the corresponding function graphs for $y_2$, $z_3$, and $z_1$ of Figure~\ref{fig:2D-CVM_Expts-scale-free-h-eq-1pt65-rev_2021-02-16_scale-free-rslts}. For visual clarity, Figure~ 6 in Maren (2021) \cite{Maren_2021_2-D-CVM-Topographies} shows only the location of these interpretation variables $y_2$, $z_3$, and $z_1$ for the original, manually-created grid, without the overlay of those same variable values for the two free energy-minimized states. 

 Figure~\ref{fig:2D-CVM_Expts-scale-free-h-eq-1pt65-rev_2021-02-16_scale-free-rslts} presents results for two cases; where \textbf{(a)} $h=1.65$, and  \textbf{(b)} $h=1.16$. The specific interpretation variable values resulting from each of these free energy minimizations are shown in the legends on the right-hand-side; those corresponding to $h=1.65$ are shown in the top-most legend (where the diamonds are each labeled ``\textbf{a}''), and those for $h=1.16$ are on the bottom (where the diamonds are each labeled ``\textbf{b}'').

These two \textit{h-values} were selected because they were deemed to be the ``best'' \textit{h-values} for modeling two different, manually-designed topographies studied in  Maren (2019b, 2021) \cite{Maren_2019_2D-CVM-FE-fundamentals-and-pragmatics, Maren_2021_2-D-CVM-Topographies}. The value of $h = 1.65$ was originally selected for modeling a highly-agglutinated initial grid, and it was used (along with $h=1.16$) to create two different models for the initial data grid shown in Figure~~\ref{fig:2D-CVM-init-and-FEMin-scale-free_h-eq-1pt165_crppd_2019-06-26}. When $h=1.65$, the resulting system was pushed to a more agglutinated result than when $h=1.16$, that is, there was a higher incidence of like-near-like nodes.

%
\section{Background:  Active Inference, Variational Bayes, and the Kullback-Leibler Divergence}
\label{subsec:background-act-inf--var-Bayes-KL-div}

%

We briefly give some context for the use of a variational approach in this work, and include the potential connection with active inference, a method developed by Karl Friston and colleagues. We describe the Kullback-Leibler divergence, as we create (in Section~\ref{sec:methods}, ``Methods'') a new divergence measure that takes inspiration from the Kullback-Leibler approach.

%
\subsection{Background:  Variational Bayes}
\label{subsec::background-var-Bayes}
%

Variational methods center on the notion of finding the set of parameters $\{\theta\}$ that yield the ``best'' model for a given data set. In many discussions of variational methods, these models come from the family of exponential equations. See, e.g.,  Beal (2003) \cite{Beal_2003_Variational-algorithm-approx-Bayes-inference},  Wainwright and Jordan (2008) \cite{Wainwright-and-Jordan_2008_Graph-models-exp-fam-var-inf}, and  Blei et al. (2017) \cite{Blei-et-al_2017_Variational-Bayes}, and also a recent review by Zhang et al. (2019) \cite{Zhang-et-al_2019_Adv-var-Bayes}. 

Although we might consider using the variational Bayes method as a starting point for determining which model (and its associated enthalpy parameters) offers the best match for each distinct topography, we cannot do a simplistic application of variational Bayes to the 2-D CVM. Nevertheless,  Maren (2019a) \cite{ Maren_2019_Deriv-var-Bayes} suggests that if there were a means to identify the $(\varepsilon_0, h)$ parameters identifying a ``best fit'' between a 2-D CVM topographic model and an actual 2-D topographic data representation, this would allow for a new model to be used in variational applications. 

The reason that simple application of variational Bayes is unsuitable is that this method uses a divergence measure originally proposed by Kullback and Leibler (1951) \cite{Kullback-Leibler-1951-Info-and-sufficiency}, which involves the logarithm of the ratio of two values. The numerator in this ratio is the probability of occurrence for each of the different observable states for the system being modeled, and the denominator is the corresponding actual fractional occurrence of that state for the model itself. 

If the ratio takes a unitary value, the logarithm goes to zero, and the Kullback-Leibler divergence goes to zero. 

The problem, as illustrated in the (admittedly toy) example used here, is that the terrain images, and the representations of the terrain images used for modeling, are constructed so that the appearance of ``black'' (``on,'' or state \textbf{A}) nodes is equal to the number of ``white'' (``off,'' or state \textbf{B}) nodes. (Again, the terms ``nodes'' and ``units'' are used interchangeably throughout.)

Further, for this study, we restrict ourselves to models where the same equiprobability condition holds. (This means that we restrict ourselves to models where $\varepsilon_0 = 0$, which indeed corresponds to the equiprobability in the initiating patterns.)

Thus, since the probability of occurrence of nodes in state \textbf{A}  is the same as for those in state \textbf{B}, in both the data representation and the model, the Kullback-Leibler divergence would be zero for any parameters tested in the model. 

This means that we need to take a step beyond the basic Kullback-Leibler divergence method, and propose a divergence suitable for working with models where the configuration variables (nearest-neighbor, next-nearest-neighbor, and triplet values) may be different in the model from those in the representation, even when the probability of occurrence of unitary nodes in a specific state is the same, or $x_1 = x_2 = 0.5$.

%
\subsection{Background: Active Inference}
\label{subsec::background-act-inf}
%

\begin{figure}[!t]
\centering
\includegraphics[width=2.5in]{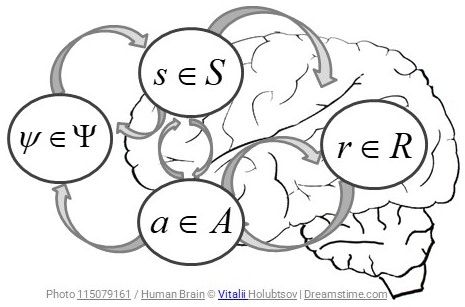}
\caption{Active inference. }  
\label{fig:Fristons-notation_2021-07-27_crppd}
\end{figure}

The notion of representing a data set via a 2-D image is relevant when we consider the \textit{active inference} methods proposed by Friston and colleagues (2013, 2015)  \cite{Friston_2013_Life-as-we-know-it, Friston-et-al_2015_Knowing-ones-place-free-energy-pattern-recognition}. See also an excellent review by Sajid et al. (2020) \cite{Sajid-and-Friston-et-al_2020_Active-Inference}. In active inference,  a representational system $R$ is constructed to represent (but not yet model) an external, observed system $\Psi$, as shown in Figure~\ref{fig:Fristons-notation_2021-07-27_crppd}.

See a detailed discussion of notation in Subsection~\ref{subsec:notation-kullback-leibler-divergence}.

%
\subsection{Background: The Kullback-Leibler Divergence}
\label{subsec:background-the-kullback-leibler-divergence}
%

In the introductory remarks, we stated that our goal is to determine which \textit{h-value} provides the ``most suitable'' resulting model for a given initial, natural topography - or, more specifically, model a \textit{representation} of a given natural topography. We have not yet defined the term ``most suitable.'' 

Naturally, our thoughts first turn to the well-known Kullback-Leibler divergence, commonly referred to as the ``K-L divergence.'' 

As we will quickly see, this divergence measure can provide us with inspiration, but is not in itself suitable for our work. 

However, before we address the kind of divergence that would indeed suit our purposes, it is helpful to look at the notation commonly used for the K-L divergence. The reason is that, in this work (and in predecessor works, and in our desires for subsequent works), we adopt the notation used by Friston and colleagues. 

Because this discussion of notation for the Kullback-Leibler divergence is subtle and nuanced, depending on the researchers whose works are used as a reference, we defer this to Subsection~\ref{subsec:notation-kullback-leibler-divergence}.

%
\section{Background: The Cluster Variation Method}
\label{subsec:background-cvm}
%

The cluster variation method (CVM), originally introduced by Kikuchi \cite{Kikuchi_1951_Theory-coop-phenomena} and later refined by Kikuchi and Brush \cite{Kikuchi-Brush_1967_Improv-CVM}, is a means for expressing the free energy of a system that expands the entropy term to include, not just the relative fractions of nodes in an ``on'' (\textbf{A}, shown as black nodes) or ``off'' (\textbf{B}, shown as white nodes) state, but also the relative proportions of nearest neighbor, next-nearest neighbor, and triplet patterns. These are collectively known as the \textit{configuration variables}.

In this section, we briefly overview  prior work using the CVM, with particular attention to works published subsequent to the first works in this series by Maren (2014, 2016) \cite{AJMaren-TR2014-003, Maren_2016_CVM-primer-neurosci}. 

For the benefit of readers who are not deeply familiar with the 2-D CVM, Section~\ref{sec:config-variables} briefly overviews the configuration variables used by Kikuchi and Brush, and in this work. The following Section~\ref{sec:2-D-CVM} reviews the 2-D CVM method itself, presenting the essential free energy equation, together with the analytic solution available when there is an equiprobable distribution of nodes into states \textbf{A} and \textbf{B}. 

The primary applications of the CVM have, up until recently, predominantly been to computations of alloy phase diagrams as well as to studies of phase transitions in alloys. This application 
area was particularly dominant in the 1980's and `90's. For example, Sanchez et al. (1984) addressed the CVM's role in describing configurational thermodynamics (and specifically phase stability) of alloys \cite{Sanchez-et-al_1984_Gen-Cluster-Descrp-Multicomp-Systems}, with a focus on the phenomenological and first principles theories of phase equilibrium. Mohri has applied the 3-D CVM to studies of alloys (e.g., Mohri, 2013 \cite{Mohri_2013_CVM}). 

Practical applications, other than to metallurgy, have remained few up until now. Albers et al. used the CVM to study efficient linkage analysis on extended pedigrees (2006) \cite{Albers-et-al_2006_CVM-efficient-linkage-analysis}, and Barton and Cocco (2013) used the CVM method (which they described as ``selective cluster expansion'') to characterize neural structural and coding properties \cite{Barton-Cocco_2013_Ising-models-neural-activity}.
 
Not included in previous literature reviews, Buzano et al. (1996) studied the 2D CVM in plaquette model form \cite{Buzano-Evangelistaand-Pelizzola_1996_Ising-models-neural-activity}, finding several distinct phases.  
 
 More recently, Sajid et al. (2021) \cite{Sajid-Convertino-and-Friston_2021_Cancer-niches-and-Kikuchi-FE} applied the 2-D CVM to the evolution of cancer niches. Specifically, they studied characterized cancer niche construction as a direct consequence of interactions between clusters of cancer and healthy cells.

The 1-D CVM is much simpler to study than the 2-D CVM, as its analytic solution is significantly easier to compute. Maren initiated this current line of work with a 1-D CVM model in 2016
 \cite{AJMaren-TR2014-003, Maren_2016_CVM-primer-neurosci}, building on prior work by Maren et al. (1984) using a linear collection of 1-D CVMs to model phase transitions and hysteresis in a solid-state oxide \cite{Maren-et-al_1984_Theoretical-model-hysteresis-solid-state-phase-trans}. This became the foundation for a neural network architecture known as the CORTECON (for Content-Retentive TEmporally-CONnected network) in 1992 \cite{Maren-Schwartz-Seyfried_1992_Config-entropy-stabilizes, Maren_1993_Free-energy-as-driving-function, Schwartz-Maren_1994_Domains-interacting-neurons}, and continued in 2015 \cite{AJMaren-HSzu-New-EEG-Measure-SPIE-STA-2015}. Since 2016, the focus has shifted to work with the 2-D CVM. 

The primary investigations into the 2-D CVM, over the past few decades, have focused on its critical behavior. Maren (2021) \cite{Maren_2021_2-D-CVM-Topographies} presented a substantive review of those studies. 

Beginning in the early 2000's, a few authors addressing general methods for machine learning and artificial intelligence included the CVM in their comprehensive reviews of methods. For example, Pelizzola broadened his earlier (1994) \cite{Pelizzola_1994_CVM-Pade-approx-crit-behavior} work to include the CVM in a more general treatment of probabilistic graph models (2005) \cite{Pelizzola_2005_CVM-stat-phys-prob-graph-models}. Similarly, Yedidia et al. described the role of the CVM as one method for belief propagation (2002) \cite{Yedidia-Freeman-Weiss_2002_Understanding-belief-prop}. Wainwright and Jordan included the CVM in their extensive monograph on graphical models, exponential families, and variational inference (2008) \cite{Wainwright-and-Jordan_2008_Graph-models-exp-fam-var-inf}. However, in all of these treatments, the CVM approach was included largely for completeness, and not as a primary method.
 
In short, there has been some applications work, and some theoretical consideration, over the past few decades. With the exception of the work by Sajid et al. (2021) \cite{Sajid-Convertino-and-Friston_2021_Cancer-niches-and-Kikuchi-FE}, there does not seem to have been much attention to the actual topographies produced during CVM free energy minimization. This is possibly because, although the analytic solution for the 2-D CVM (with staggered or zigzag chains) was produced by Kikuchi and Brush in 1967, the details of this solution were not largely known - and the derivation was complex. 

Maren presented the derivation for this analytic solution in 2014 \cite{AJMaren-TR2014-003, Maren_2016_CVM-primer-neurosci}, and followed with the derivation included as an appendix in Maren (2019b) \cite{Maren_2019_2D-CVM-FE-fundamentals-and-pragmatics}. 

The code for actually counting the configuration variables is also somewhat complex. That code is also now publicly available, and is included within the GitHub repository associated with this work \cite{AJM-Github-2-D-CVM-w-Var-Bayes-2022}. 

Thus, it is reasonable that both theoretical investigations as well as practical applications of the 2-D CVM can gain rapid momentum within the near-term.

%
\section{The Configuration Variables}
\label{sec:config-variables}
%

This section on the \textit{configuration variables} briefly summarizes material presented in Section 2.1 as well as Appendices A.1 - A.4 of Maren (2021) \cite{Maren_2021_2-D-CVM-Topographies}, and in greater detail in Sections 3 and 4 of Maren (2019b) \cite{Maren_2019_2D-CVM-FE-fundamentals-and-pragmatics}.

The~topographic characteristics in the case of a simple Ising system are not~important. In contrast, when we work with a 2-D CVM, the topographies are visible, and are characterized by the \textit{configuration variables}. 

The configuration variables are denoted as: 

\begin{itemize}
\setlength{\itemsep}{1pt}
\item $x_i$ - Single units, 
\item $y_i$ - Nearest-neighbor pairs, 
\item $w_i$ - Next-nearest-neighbor pairs, and 
\item $z_i$ - Triplets. 
\end{itemize}

The instances for these configuration variables are summarized in Table~\ref{tbl:config-variables-table}.

\begin{table}
    \caption{Configuration Variables for the Cluster Variation Method}
    \label{tbl:config-variables-table}
    \centering
    \vspace{3mm}
    \begin{tabular}{|p{5cm}|p{2cm}|p{2cm}|}
    \hline
	 \multicolumn{1}{|>{\centering\arraybackslash}m{5cm}|}	{\textbf{Name}} 
    & 	 \multicolumn{1}{|>{\centering\arraybackslash}m{2cm}|}	{\textbf{Variable}}  
    &   \multicolumn{1}{|>{\centering\arraybackslash}m{2cm}|}   {\textbf{Instances}} \T\B \\ 
    \hline    	    

	 \multicolumn{1}{|>{\centering\arraybackslash}m{5cm}|}	{Unit} 
    &   \multicolumn{1}{|>{\centering\arraybackslash}m{2cm}|}	{$x_i$} 	
    &   \multicolumn{1}{|>{\centering\arraybackslash}m{2cm}|}	{$2$} \\ [3pt] 			  
  
	 \multicolumn{1}{|>{\centering\arraybackslash}m{5cm}|}	{Nearest-neighbor} 
    &   \multicolumn{1}{|>{\centering\arraybackslash}m{2cm}|}	{$y_i$} 	
    &   \multicolumn{1}{|>{\centering\arraybackslash}m{2cm}|}	{$3$} \\ [3pt] 	    

	 \multicolumn{1}{|>{\centering\arraybackslash}m{5cm}|}	{Next-nearest-neighbor} 
    &   \multicolumn{1}{|>{\centering\arraybackslash}m{2cm}|}	{$w_i$} 	
    &   \multicolumn{1}{|>{\centering\arraybackslash}m{2cm}|}	{$3$} \\ [3pt] 	
    
	 \multicolumn{1}{|>{\centering\arraybackslash}m{5cm}|}	{Triplet } 
    &   \multicolumn{1}{|>{\centering\arraybackslash}m{2cm}|}	{$z_i$} 	
    &   \multicolumn{1}{|>{\centering\arraybackslash}m{2cm}|}	{$6$} \\ [5pt] 	    
 
    \hline
  \end{tabular}
\end{table}

\begin{figure}[!t]
\centering
\includegraphics[width=3in]{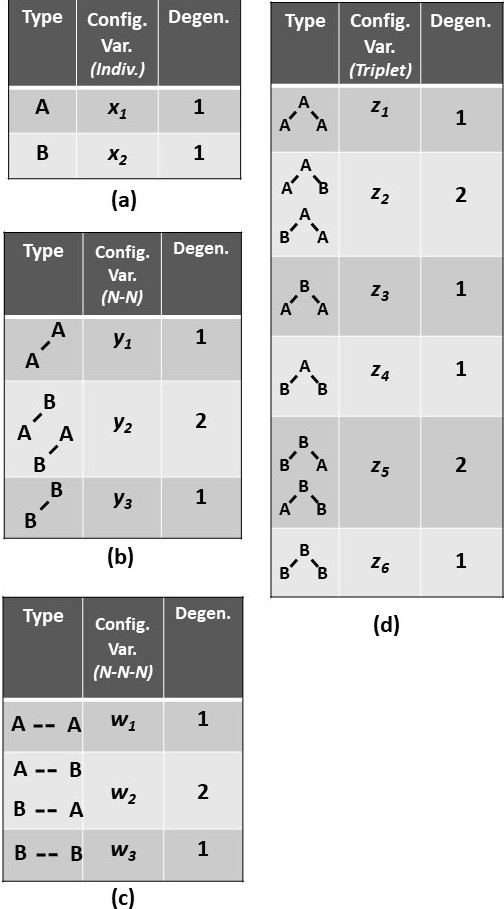}
\caption{Illustration of the configuration variables for the 2-D CVM, created using overlapping zigzag chains: (a) the fractions in the two activation states; (b) the nearest-neighbor fractions (read as diagonals in the grid layouts), (c) the next-nearest-neighbor fractions, and (d) the six triplet configuration variable types. }  
\label{fig:Config-var-weights_v4_2021-07-13-all-z_crppd}
\end{figure}

Figure~\ref{fig:Config-var-weights_v4_2021-07-13-all-z_crppd} illustrates these configuration variables, and shows how four of them appear in two different ways, so that they have a degeneracy factor of two.

For a bistate system (one in which the units can be in either state \textbf{A} or state \textbf{B}), there are six different ways in which the triplet configuration variables ($z_i$) can be constructed, as shown in Figure~\ref{fig:Config-var-weights_v4_2021-07-13-all-z_crppd}. 

Notice that within Figure~\ref{fig:Config-var-weights_v4_2021-07-13-all-z_crppd}, the triplets $z_2$ and $z_5$ have two possible configurations each: \textbf{A}-\textbf{A}-\textbf{B} and \textbf{B}-\textbf{A}-\textbf{A} for $z_2$, and \textbf{B}-\textbf{B}-\textbf{A} and \textbf{A}-\textbf{B}-\textbf{B} for $z_5$. This means that \textbf{\textit{there is a degeneracy factor of 2}} for each of the $z_2$ and $z_5$ triplets. We denote the degeneracy factor for the $z_i$ units as $\gamma_i$, where $\gamma_2 = \gamma_5 = 2$, and $\gamma_i = 1$ for all other $z_i$. 

Similarly, there is a degeneracy factor $\beta_2 = 2$  for the pairwise combinations $y_2$ and $w_2$,  as $y_2$ and $w_2$ can each be constructed as either \textbf{A}-\textbf{B} or as \textbf{B}-\textbf{A} for $y_2$, or as \textbf{B}-~-\textbf{A} or as \textbf{A}-~-\textbf{B} for $w_2$. 

The reader who is unfamiliar with the intricacies of the 2-D CVM is gently encouraged to read Section 3, ``The Configuration Variables,'' and Section 4, ``Interpreting Configuration Variables,'' of Maren (2019b) \cite{Maren_2019_2D-CVM-FE-fundamentals-and-pragmatics}. This work takes a largely tutorial approach and contains illustrative details not found in other published works.

%
\section{The 2-D Cluster Variation Method}
\label{sec:2-D-CVM}
%

This section on the 2-D CVM free energy, and the influence of parameter values on the free energy, largely duplicates material presented in Section 2.3 of Maren (2021) \cite{Maren_2021_2-D-CVM-Topographies}, and also (in more detail) in Section 5 of Maren (2019b) \cite{Maren_2019_2D-CVM-FE-fundamentals-and-pragmatics}. The material is presented here as a courtesy to the reader, who may not have read the predecessor publications. 

The essential notion of the CVM is that we work with a more complex expression for the free energy in a system. 

As a point of comparison, the basic Ising equation is

\begin{equation}
\label{Bar-F-basic-Ising-basic-eqn}
\bar{F} = F/(Nk_{\beta}T) =  \bar{H}  - \bar{S},
\end{equation}

\noindent
where $F$ is the free energy, $H$ is the enthalpy and $S$ is the entropy for the system, and where $N$ is the total number of units in the system, $k_\beta$ is Boltzmann's constant, and $T$ is the temperature. 

For working with abstract systems, the total $Nk_{\beta}T$ can be absorbed into a \textit{reduced energy formalism}, as these values are constants during system operations. This leads to the \textit{reduced representations} of $\bar{F}$, $\bar{H}$, and $\bar{S}$. We will work consistently with reduced representations throughout this work. 

In a simple Ising model, both the reduced enthalpy $\bar{H}$ and the reduced entropy  $\bar{S}$ can be computed based on only the relative fraction of active units in a bistate system. That is, there are only two kinds of computational units; \textit{active} ones in state \textbf{A}, where the fraction of these units is denoted $x_1$, and \textit{inactive} ones in state \textbf{B}, where the fraction of these units is denoted $x_2$. (Of course, $x_1 + x_2 = 1.0$.)

In contrast to the simple entropy used in the basic Ising model in statistical mechanics, in the CVM approach, we expand the entropy term. The CVM entropy term considers not only the relative fractions of units in states \textbf{A} and \textbf{B}, but also terms associated with the \textit{configuration variables}, as described earlier in the preceding Section~\ref{sec:config-variables}.

%
\subsection{The 2-D CVM Free Energy}
\label{subsec:2D-CVM-free-energy}
%

We can write the 2-D CVM free energy, using the formalism first introduced by Kikuchi in 1951 \cite{Kikuchi_1951_Theory-coop-phenomena}, and then further advanced by Kikuchi and Brush (1967) \cite{Kikuchi-Brush_1967_Improv-CVM} (explicitly for the 2-D CVM), as

\begin{equation}
\label{eqn:Bar-F-2-D-basic-eqn}
  \begin{aligned}
\bar{F}_{2-D} = F_{2-D}/N = \\
  & \bar{H}_{2-D}- \bar{S}_{2-D} \\
+ & \mu (1-\sum\limits_{i=1}^6 \gamma_i  z_i )+4 
\lambda (z_3+z_5-z_2-z_4),
  \end{aligned}
\end{equation}

\noindent
where $\mu$ and $\lambda$ are Lagrange multipliers, and we have set $k_{\beta}T = 1$. 

In the following two subsections, we separately address the enthalpy and the entropy terms for the 2-D CVM. 

The material contained in the following subsections is extracted from the prior presentation found in  (Maren (2019b) \cite{Maren_2019_2D-CVM-FE-fundamentals-and-pragmatics}). It is presented here to provide the reader with a continuous overview of the 2-D CVM. 

%
\subsection{The 2-D CVM Enthalpy}
\label{subsec:2D-CVM-enthalpy}
%

The enthalpy in a simple Ising system is traditionally given as 

\begin{equation}
\label{eqn:Bar-H-2-D-basic-eqn}
  \begin{aligned}
	\bar{H}_{2-D} = H_{2-D}/N \\
  & = \bar{H}_0 + \bar{H}_1 \\
  & = \varepsilon_0 x_1 + c x_1^2,  \\
  \end{aligned}
\end{equation}

\noindent
where $H_0$ and $c$ are constants.

The first term on the RHS (Right-Hand-Side) corresponds to the \textit{activation enthalpy}, or enthalpy associated with each active unit. The second term on the RHS corresponds to the \textit{interaction enthalpy}, or energy associated with pairwise interactions between active units. 

In contrast, the enthalpy for the 2-D CVM is given as 

\begin{equation}
\label{eqn:Bar-H-2-D-basic-eqn}
  \begin{aligned}
	\bar{H}_{2-D} = H_{2-D}/N \\
  & = \bar{H}_0 + \bar{H}_1 \\
  & = \varepsilon_0 x_1 + \varepsilon_1(-z_1+z_3+z_4-z_6) 
  \end{aligned}
\end{equation}

We will separately consider these two terms; the first pertaining to the activation enthalpy, and the second to the interaction enthalpy. 

%
\subsubsection{The 2-D CVM Activation Enthalpy (Role of $\varepsilon_0$)}
\label{subsubsec:2D-CVM-activ-enthalpy-eps0}
%

In the original work by Kikuchi and Brush, Eqn.~\ref{eqn:Bar-H-2-D-basic-eqn} is simplified (K\&B Eqns. I.16 and I.17) to 

\begin{equation}
\label{eqn:Bar-H-2-D-basic-eqn-Kikuchi-and-Brush}
\bar{H}_{2-D} = \varepsilon_1(-z_1+z_3+z_4-z_6), 
\end{equation}

\noindent
that is, they omit the term linear in $x_1$; the activation enthalpy. 

This simplification allowed Kikuchi and Brush to provide an analytic solution for the 2-D CVM free energy, for the specific case where there is an equiprobable distribution of units, that is, $x_1 = x_2 = 0.5$. This equiprobable distribution is achieved only when the activation enthalpy $\varepsilon_0 = 0$. 

When this equiprobable distribution case holds, then there are many other simplifications possible for the configuration variables, e.g. $z_1 = z_6$, etc. This reduces the total number of configuration variables that are brought into play, so that the analytic solution for the free energy becomes possible. (This is discussed in Subsection~\ref{subsec:free-energy-analytic-solution}.)

When the activation enthalpy $\varepsilon_0 > 0$, there will no longer be an equiprobable distribution of nodes into states \textbf{A} and \textbf{B}. Instead, the units in state \textbf{A} have an energy associated with them that is greater than that of the units in state \textbf{B}. Thus, an equilibrium solution will favor having fewer units in state \textbf{A}.

There is no analytic solution for this case, other than that in which the interaction enthalpy is zero ($\varepsilon_1 = 0$). This latter case is trivial to solve, and is not particularly interesting, as the distribution of different kinds of nearest-neighbor pairs and triplets will be probabilistically random, excepting only that the proportions of units in states \textbf{A} and \textbf{B}, respectively, will be skewed by the activation enthalpy parameter $\varepsilon_0$. 

However, this case where $\varepsilon_0 > 0$ and $\varepsilon_1 = 0$ ($h = 1$) provides the other axis for the $(\varepsilon_0, h)$ phase space, and is thus worth expressing. The details are presented in (Maren (2019b) \cite{Maren_2019_2D-CVM-FE-fundamentals-and-pragmatics}).

%
\subsubsection{The 2-D CVM Interaction Enthalpy (Role of $\varepsilon_1$)}
\label{subsubsec:2D-CVM-interact-enthalpy-eps1}
%

As previously noted, the typical expression for the interaction enthalpy is a quadratic term in $x_1$, that is, $H_1 = c x_1^2$. The parameter $c$ encompasses both the actual interaction energy for each pairwise interaction, and a constant that expresses the distribution of pairwise interactions as a simple linear function of the fraction of active units ($x_1$) surrounding a given active unit. This is then multiplied by the total fraction of active units, giving the quadratic expression. 

In the expression for the 2-D CVM interaction enthalpy, we have terms that expressly identify the total fraction of nearest-neighbor ``unlike'' pairs ($y_2$) and ``like'' pairs ($y_1$ and $y_3$). Thus, we can replace $c x_1^2$ with the fraction of ``unlike`` pairs (counted twice, to account for the degeneracy in how these pairs can be counted), and the fractions of ``like'' nearest neighbor pairs. 

We recognize that this is a simplification; we are not counting interaction energies due to next-nearest neighbor pairs, the $w_i$, nor from the triplets $z_i$. We are, effectively, subsuming these into the pairwise interactions that are being modeled with the $y_i$. 

We take the interaction enthalpy parameter $\varepsilon_1$ to be a positive constant. 

We envision a system in which the free energy is reduced by creating nearest-neighbors of like units, that is, \textbf{A}-\textbf{A} or \textbf{B}-\textbf{B} pairs. (Decreasing the interaction enthalpy leads to decreasing the free energy, which is desired as we go to a free energy minimum, or equilibrium state.) Similarly, the interaction enthalpy should increase with unlike pairs, or \textbf{A}-\textbf{B} pairs (or vice versa). 

Further following the approach introduced by Kikuchi and Brush, we can use the equivalence relations, described in Appendix~\ref{subsec:App-A-Equiv-Rel_Config-Vars}, to rewrite Eqn.~\ref{eqn:Bar-H-2-D-basic-eqn-Kikuchi-and-Brush} in terms of the $y_i$ variables instead of the $z_i$, as

\begin{equation}
\label{eqn:enthalpy-Kikuchi-Brush}
  \begin{aligned}
\bar{H}_{2-D} = H_{2-D}/N = 
\varepsilon_1(2y_2 - y_1 - y_3). 
  \end{aligned}
\end{equation}

We interpret this equation by noting that as we increase the fraction of unlike unit pairings (\textbf{A}-\textbf{B} or \textbf{B}-\textbf{A} pairs, expressed using $y_2$), we raise the interaction enthalpy. At the same time, if we're increasing \textit{unlike} unit pairings, we are also decreasing \textit{like} unit pairings (\textbf{A}-\textbf{A} and \textbf{B}-\textbf{B}, or $y_1$ and $y_3$, respectively), so that we are again increasing the overall interaction energy. (Note that the \textit{like} unit pairings show up in Eqn.~\ref{eqn:enthalpy-Kikuchi-Brush} with a negative sign in front of them.)

The more that we increase $\varepsilon_1$ (increase $h$), the more that we \textit{decrease} the overall enthalpy (and thus move to a lower free energy) by putting unlike units together, that is, by increasing $y_2$. 

We can anticipate the role of $\varepsilon_1$ (or correspondingly, $h$) in finding the ``best fits'' to a given topography by referring back to the three natural topographies shown in Figure~\ref{fig:Rocks-three-topographies-horiz-crppd-2020-11-16}. 

In this figure, we note that Figure~\ref{fig:Rocks-three-topographies-horiz-crppd-2020-11-16}\textbf{(a)} (on the LHS) portrays highly-contiguous areas, whereas Figure~\ref{fig:Rocks-three-topographies-horiz-crppd-2020-11-16}\textbf{(c)} shows a much more granular terrain, and Figure~\ref{fig:Rocks-three-topographies-horiz-crppd-2020-11-16}\textbf{(b)} is intermediate. 

Thus, we would expect that the ``best fit'' to Figure~\ref{fig:Rocks-three-topographies-horiz-crppd-2020-11-16}\textbf{(a)} would be provided with a relatively high $\varepsilon_1$ (higher $h$) value, whereas the ``best fit'' to  Figure~\ref{fig:Rocks-three-topographies-horiz-crppd-2020-11-16}\textbf{(c)} would be with a much smaller $\varepsilon_1$ (or where $h$ is closer to 1). We would anticipate that the $\varepsilon_1$ providing a ``best fit'' to Figure~\ref{fig:Rocks-three-topographies-horiz-crppd-2020-11-16}\textbf{(b)} would be between these two other $\varepsilon_1$  values.

%
\subsection{The 2-D CVM Entropy}
\label{subsec:2D-CVM-entropy}
%

In a simple Ising model, the entropy $S$ can be computed based on only the relative fraction of active units in a bistate system. That is, there are only two kinds of units; \textit{active} ones in state \textbf{A}, where the fraction of these units is denoted $x_1$, and \textit{inactive} ones in state \textbf{B}, where the fraction of these units is denoted $x_2$. (Of course, $x_1 + x_2 = 1.0$.)

In contrast to the simple entropy used in the basic Ising model, in the CVM approach, we expand the entropy term. The CVM entropy term considers not only the relative fractions of units in states \textbf{A} and \textbf{A}, but also the set of \textit{configuration variables}, as described in Section~\ref{sec:config-variables}.

Our entropy term now involves what are essentially \textit{topographic} variables. That is, the location of one unit in conjunction with another now makes a substantial difference. 

The reduced entropy for the 2-D CVM is given as

\begin{equation}
\label{eqn:Bar-S-2-D-basic-eqn}
  \begin{aligned}
\bar{S}_{2-D} = S_{2-D}/N = \\
 & 2 \sum\limits_{i=1}^3 \beta_i Lf(y_i))
          + \sum\limits_{i=1}^3 \beta_i Lf(w_i) \\
 &      - \sum\limits_{i=1}^2 \beta_i Lf(x_i)
          - 2 \sum\limits_{i=1}^6 \gamma_i Lf(z_i), \\
  \end{aligned}
\end{equation}

\noindent
where $Lf(v)=vln(v)-v$.

A more detailed discussion of the entropy term is given in Kikuchi and Brush (1967) \cite{Kikuchi-Brush_1967_Improv-CVM} and also in Maren (2019b) \cite{Maren_2019_2D-CVM-FE-fundamentals-and-pragmatics}).

%
\subsection{Free Energy Analytic Solution}
\label{subsec:free-energy-analytic-solution}
%

Kikuchi and Brush (1967) \cite{Kikuchi-Brush_1967_Improv-CVM} provided the results of an analytic solution for the 2-D CVM free energy, for the specific case where $x_1 = x_2 = 0.5$, which represents an equiprobable distribution between states \textbf{A} and \textbf{B}. 

Specifically, making certain assumptions about the Lagrange multipliers shown in Eqn.~\ref{eqn:Bar-F-2-D-basic-eqn}, we can express each of the configuration variables in terms of $\varepsilon_1$. 

More usefully, since the expression actually involves the term $exp(2\varepsilon_1)$, and not $\varepsilon_1$ itself, it is much easier to use the substitution variable $h = exp(2\varepsilon_1)$. We refer to $h$ (or sometimes, the \textit{h-value}), as the \textit{interaction enthalpy parameter} throughout. 

The full derivation of the set of equations giving the configuration variable values at equilibrium (i.e., at $x_1 = x_2 = 0.5$) was originally presented in Maren (2014) \cite{AJMaren-TR2014-003}, and, more recently, given in Appendix A of (Maren (2019b) \cite{Maren_2019_2D-CVM-FE-fundamentals-and-pragmatics}).

The full set of these analytic expressions for the various configuration variables is couched in terms of a denominator involving \textit{h}, specifically 

\begin{equation}
\Delta =-h^2 +6h - 1,
\label{Kikuchi-and-Brush-Delta-def-Eqn-Ipt24-main-text}
\end{equation}

\noindent
which Kikuchi and Brush present as their Eqn. (I.24) \cite{Kikuchi-Brush_1967_Improv-CVM}. 

We recall that at the equiprobable distribution point, where $x_1 = x_2$, we have a number of other equivalence relations, e.g. $z_1 = z_6$, etc. 

We then (following Kikuchi and Brush, in their Eqn.~(I.25)) identify each of the remaining configuration variables as

\begin{equation}
\begin{array}{lll}
  y_1 = y_3 &=&  \frac {( 3h  - 1)} { 2 \Delta}\\
  y_2  &=&  \frac {h( -h  +  3) } { 2 \Delta}\\
  w_1 =w_3 &=& \frac {( h  +  1)^2 } { 4 \Delta}\\ 
  w_2  &=&  \frac {(3h-1)( -h  +  3) } { 4 \Delta}\\
  z_1 = z_6 &=& \frac {(3h-1)( h  +  1) } { 8 \Delta}\\
  z_2 = z_5 &=& \frac {(3h-1)( -h  +  3) } { 8 \Delta}\\ 
  z_3 = z_4 &=&  \frac {( -h  +  3) ( h  +1)} { 8 \Delta}\\ 
 \end{array}
 \label{config-var-analytic-eqns-K-and-B-main-text}
\end{equation}

%
\subsection{Divergence in the analytic solution }
\label{subsec:divergence-analytic}
%

As is obvious from Eqn.~\ref{Kikuchi-and-Brush-Delta-def-Eqn-Ipt24-main-text}, there will be a divergence in the analytic solution for the configuration variables at the free energy minimum, because the analytic solution contains a denominator term that is quadratic in \textit{h}. Specifically, the term diverges for $h = 0.172$ or $h = 5.828$. 

We are interested in the latter case, where the value of $h>1$ indicates that $\varepsilon_1 >0$, which is the case where the interaction enthalpy favors like-near-like interactions, or some degree of gathering of similar units into clusters. 

As reported in Maren (2019b, 2021) \cite{Maren_2019_2D-CVM-FE-fundamentals-and-pragmatics, Maren_2021_2-D-CVM-Topographies}, when the \textit{h-value} becomes large (e.g., $h > 2$), then the enthalpy term dominates the entropy term, and useful free energy minima are not readily found. Thus, there is a practical limit on the useful range of \textit{h-values}; approximately $1 \leq h \leq 2$. 

In the following study, we find that \textit{h-values} within this range have been appropriate for modeling the naturally-occurring topographies used to generate our data representations.

%
\section{Data}
\label{sec:data}
%

The initial data chosen for this experiment was the set of three natural terrain images, as shown in Figure~\ref{fig:Rocks-three-topographies-horiz-crppd-2020-11-16}.

These images were selected because they were composed of:

\begin{itemize}
\setlength{\itemsep}{1pt}
\item \textbf{\textit{Naturally-occurring terrain}}, which might be at or nearly at an equilibrium pattern distribution in each case, 
\item \textbf{\textit{Largely black-and-white elements}}, which might be suitably be represented as a bistate system, and where the distribution of ``on'' (black) and ``off'' (white) nodes would potentially be close to equiprobable, and 
\item \textbf{\textit{Different distributions of configuration variables}}  (across the three images), which should yield correspondingly different \textit{h-values} for the representation corresponding to each image.
\end{itemize}

%
\subsection{Data Processing Overview}
\label{subsec:data-proc-overview}
%

Following both the notions and notation used by Friston and colleagues \cite{Friston_2013_Life-as-we-know-it, Friston-et-al_2015_Knowing-ones-place-free-energy-pattern-recognition}, we treat the actual data itself as the system $\Omega$, and construct a representation of that system $R$. We apply the methods described here to model the representation, not the actual data itself. This suggests that one aspect of an evolving research approach will be to refine how the representation $R$ is built from the original data $\Omega$. 

As a preliminary step, we constructed a 2-D CVM overlay grid template with ``staggered'' nodes, as was shown previously in Figure~\ref{fig:2D-CVM-init-and-FEMin-scale-free_h-eq-1pt165_crppd_2019-06-26}. 

This grid design was constructed with three purposes in mind:

\begin{itemize}
\setlength{\itemsep}{1pt}
\item \textbf{\textit{Scaled to fit easily against the image}}, so that it could not only be laid against each terrain image but its position could be adjusted, 
\item \textbf{\textit{Small enough to allow for individual assessment of each node}}; since this is a preliminary investigation, and the focus is on both  evolving and testing a new method, and also demonstrating how it is employed, it was important that the representation for each initial image was small enough to allow black/white decisions to be made individually for each node in the grid mask, and also to allow for judgment calls so that the resulting node set was equiprobably black and white (``on'' and ``off,'' respectively), and 
\item \textbf{\textit{Large enough to capture the different distributions of configuration variables}}, so that we would clearly have three different \textit{h-values} found for the representation corresponding to each image.
\end{itemize}

All of the data used, the computer programs, and the experimental results are available in a public GitHub repository \cite{AJM-Github-2-D-CVM-w-Var-Bayes-2022}. 
 
%
\subsection{An Unusual Data Preparation Challenge}
\label{subsec:unusual-challenge}
%

One of the most delicate and challenging aspects of data preparation was ensuring that the number of artifically-induced changes to the configuration variables was minimized. 

As a particular example, Figure~\ref{fig:Rocks-three-topographies-horiz-crppd-2020-11-16}\textbf{(a)} shows relatively large masses of black and white nodes. If a grid were simply placed over this image, and the usual wrap-around method (used in Maren (2019b and 2021 \cite{Maren_2019_2D-CVM-FE-fundamentals-and-pragmatics, Maren_2021_2-D-CVM-Topographies}) was used to create a complete torus ``envelope'' (both horizontally and vertically), then it is likely that an unusually high number of like-vs-unlike nearest neighbors $y_2$ would be introduced. The values for the other configuration variables would be similarly influenced. 

This artifact would be unavoidable in the case of Figure~\ref{fig:Rocks-three-topographies-horiz-crppd-2020-11-16}\textbf{(a)}, since the upper and right-most units were largely black (``on'') and the lower and left-most units were largely white (``off''). 

Thus, the grid that was actually used was cut back to one-quarter of the original size. This grid was positioned in the near-center of the image, capturing an (approximate) even number of black and white nodes. 

The ``on''/''off'' values for the nodes in this first step of constructing the representation were noted, and some (very few) adjustments were made in interpreting certain grey-valued nodes so that the resulting numbers of ``on'' and ``off'' nodes were precisely equal. 

Then, the node's values were extended (or wrapped around) to both the right and left sides, where the numbers of nodes added to the right and left were each, respectively, half the width of the grid overlay that was actually used. (In this case, the total grid size was an 8x8 grid, so that four nodes were added to each of the right and left sides.) This was done in a mirror-image manner, so that the $y_2$ values that would result when the ``envelope'' was joined at the horizontal edges would be the same as the  $y_2$ values from the center of the grid. This in itself introduced some artifice, but was better than creating an envelope that butted the black nodes from the right-hand-side  of the image against the white nodes on the left. (That would have introduced an abnormally high $y_2$ value.) 

Doing this step increased the total number of representation units from 8x8 to 8 (vertical) by 16 (horizontal). This step is documented in the MS Powerpoint$^{TM}$ detailed ``Data'' documentation slidedeck accompanying this work, publicly available in the corresponding GitHub repository \cite{AJM-Github-2-D-CVM-w-Var-Bayes-2022}.

Once this was accomplished, a similar mirror-image was constructed vertically, resulting in a final 16x16 unit representation. This is again illustrated in the detailed series of figures, documenting the creation of the ``data representation'' grid, in the MS Powerpoint$^{TM}$ ``Data'' documentation slidedeck stored in the GitHub repository \cite{AJM-Github-2-D-CVM-w-Var-Bayes-2022}.

%
\subsection{Obtaining Initial Configuration Variable Values}
\label{subsec:initial-config-var-values}
%

Once the data representation was created, the grid representation (the set of pattern ``on''/''off'' values) was manually entered into a Python program designed to count the 2-D CVM configuration variable values. (This program is available in the same GitHub repository in which the MS Powerpoint$^{TM}$ ``Data'' documentation slidedeck is stored \cite{AJM-Github-2-D-CVM-w-Var-Bayes-2022}). 

This program not only confirmed that there were equal numbers of ``on'' and ``off'' units ($x_1 = x_2 = 0.5$), but also yielded values for the $y_i$, $w_i$, and $z_i$ variables.

%
\subsection{Obtaining Initial Range for Testing \textit{h-values}}
\label{subsec:obtain-initial-h-value-test-range}
%

The premise for this work using natural terrain images as starting points was that each terrain would be approximately at an equilibrium value. There would, however, likely be divergence from any possible equilibrium point suggested in the data representations constructed for this study. The reasons would likely be:

\begin{itemize}
\setlength{\itemsep}{1pt}
\item \textbf{\textit{Forcing an equiprobable distribution}}; even though each image and each positioning of the representation grid on each image was done to maximize the likelihood of an equiprobable bistate unit distribution, such a distribution could not be guaranteed - particularly since a very small grid had to be used; this was necessary to construct the envelope mirror-images, 
\item \textbf{\textit{Boundary artifacts}}; even though the entire process of creating the mirror-image wrap-around was designed to minimize extreme boundary artifacts, it is inevitable that some would be introduced, and this is particularly true given the relatively large size of the boundary versus the grid interior, and 
\item \textbf{\textit{Inherent non-equilibrium configurations}}; the images obtained for this study were selected in the hope that each instance of natural terrain would be approximately at equilibrium; this is not necessarily and always the case.
\end{itemize}

These influences can be mitigated in future studies. This mitigation can be accomplished by using a more granular grid, so that the influence of boundary artifacts is minimized. Further, as we map out the phase space of configuration variables for the $(\varepsilon_0, \varepsilon_1)$ parameter pairs, and are able to work with non-zero values for $\varepsilon_0$, we will be less constrained. (This also implies that we are moving away from the analytic solution that is useful only when $\varepsilon_0 = 0$ as a reference point.)

Given that each of the initial image representations was likely to yield configuration variable values that \textit{did not} depict an equilibrium situation, our first step - once the set of configuration variable values had been obtained - was to plot those values on the 2-D CVM analytic solution graph, where such a graph was shown earlier in  Figure~\ref{fig:2D-CVM_Expts-scale-free-h-eq-1pt65-rev_2021-02-16_scale-free-rslts}.

This graph shows the analytically-derived equilibrium values of select configuration variable values (specifically, the interpretation variables $y_2$, $z_1$, and $z_3$) against a range of \textit{h-values} for the 2-D CVM. (Details are found in the MS Powerpoint$^{TM}$ ``Data''  deck stored in the associated GitHub repository \cite{AJM-Github-2-D-CVM-w-Var-Bayes-2022}.)

There is typically a range of the \textit{h-values} that correspond to the three different configuration variable values plotted. As an illustration, Table~\ref{config-var-values-highly-massed-top-rep} shows the approximate (visually-estimated) \textit{h-values} corresponding to the three configuration variables $y_2$, $z_1$, and $z_3$ associated with the highly-massed topography illustrated in Figure~\ref{fig:Rocks-three-topographies-horiz-crppd-2020-11-16}\textbf{(a)}. 

This range is used to establish an initial range of \textit{h-values} that will be tested to find the ``most suitable'' \textit{h-value} for a given image representation. For example, the potential range of \textit{h-values} for the topography representation based on the image shown in Figure~\ref{fig:Rocks-three-topographies-horiz-crppd-2020-11-16}\textbf{(a)} would be $1.44 \leq h \leq 1.54$. Naturally, a somewhat larger range was used for the tests.  

(As a spoiler alert: the actual \textit{h-value} that provided the best fit turned out to be much higher, that is, above the analytically-expected value; the best fit came when $h = 1.95$.)

\begin{table}[!t]
\renewcommand{\arraystretch}{1.3}
\caption{Configuration Variable Values for the Highly-Massed Topography Representation shown in Figure~\ref{fig:Rocks-three-topographies-horiz-crppd-2020-11-16}\textbf{(a)}. \textit{(*Note: The value given for $y_2 = 0.125$ has been divided by two from the original value of $y_2 = 0.250$ to account for the degeneracy of the $y_2$ term.)}}
\label{config-var-values-highly-massed-top-rep}
\centering
\begin{tabular}{|c||c|c|}
\hline
\textbf{Config. Var. } & \textbf{Config. Var. Value} & \textbf{Approx. Corresp. \textit{h-value}} \\
\hline 
$y_2$ & 0.125* & 1.49 \\
\hline
$z_1$ & 0.315 &  1.54 \\
\hline
$z_3$ & 0.065 & 1.44 \\
\hline
\end{tabular}
\end{table}

%
\subsection{Summary of Data Preparation Results}
\label{subsec:summary-data-prep}
%

We repeated the process described in previous subsection for the natural terrain images shown in Figure~\ref{fig:Rocks-three-topographies-horiz-crppd-2020-11-16}\textbf{(b)} and \textbf{(c)}. (Details are found in in the MS Powerpoint$^{TM}$ ``Data'' deck stored in the GitHub repository \cite{AJM-Github-2-D-CVM-w-Var-Bayes-2022}.) 

Table~\ref{h-value-range-three-image-reps} shows the low and high ends of the \textit{h-value} range for each of the three image representations. Note that these initial \textit{h-values} were suggested by the configuration variable values found for each of the three (and as it turns out, four) initial representations. The \textit{h-value} that actually yielded the lowest free energy for each case was often higher than initially anticipated; this is somewhat in line with the findings from Maren (2019b, 2021)) \cite{Maren_2019_2D-CVM-FE-fundamentals-and-pragmatics, Maren_2021_2-D-CVM-Topographies}  that the free energy-minimizing \textit{h-values} did not always correspond to the analytic solutions. See Section~\ref{sec:results}, \textit{Results}, for more details. 

While the data preprocessing method described in Subsections~\ref{subsec:data-proc-overview} - \ref{subsec:obtain-initial-h-value-test-range} was necessary for the first image (Figure~\ref{fig:Rocks-three-topographies-horiz-crppd-2020-11-16}\textbf{(a)}), it was not necessary for the more granular topographies shown in Figures~\ref{fig:Rocks-three-topographies-horiz-crppd-2020-11-16}\textbf{(b)} and \textbf{(c)}.  

Instead, for Figures~\ref{fig:Rocks-three-topographies-horiz-crppd-2020-11-16}\textbf{(b)} and \textbf{(c)}, the full 16x16 unit grid overlay was used. Additionally, for Figure~\ref{fig:Rocks-three-topographies-horiz-crppd-2020-11-16}\textbf{(c)}. a second representation grid could be drawn by shifting the overlay a few units to the left and down. This gave us two representations that would be likely to have similar \textit{h-values}.

\begin{table}[!t]
\renewcommand{\arraystretch}{1.3}
\caption{Range of \textit{h-values} Suggested by Initial Configuration Variables of Four Different Image Terrain Representations (Figure~\ref{fig:Rocks-three-topographies-horiz-crppd-2020-11-16}\textbf{(a-c)})}
\label{h-value-range-three-image-reps}
\centering
\begin{tabular}{|c||c|c|}
\hline
\textbf{Image Rep.} & \textbf{Low \textit{h-value}} & \textbf{High \textit{h-value}} \\
\hline 
\textbf{(a)} \textit{(Pattern 1)} & $1.44$ & $2.00$ \\
\hline
\textbf{(b)} \textit{(Pattern 2)} & $1.04$ & $1.36$ \\
\hline
\textbf{(c)} \textit{(Pattern 3a)} & $1.02$ & $1.18$ \\
\hline
\textbf{(c)} \textit{(Pattern 3b)} & $1.05$ & $1.16$ \\
\hline
\end{tabular}
\end{table}

%
\section{Methods}
\label{sec:methods}
%

From the previous Subsection~\ref{subsec:summary-data-prep}, we see that for each of the four natural topography representations used, we have a potential \textit{h-value} range. (Recall that of the three natural topographies illustrated in Figure~\ref{fig:Rocks-three-topographies-horiz-crppd-2020-11-16}, we were able to extract two partially overlapping representations from the third image, giving rise to a total of four different representations used for our study.)  

Our goal now is to determine which \textit{h-value} provides the ``most suitable'' resulting model, for each of the four different data representations. We expect that there will be some reasonable progression across the four different \textit{h-values} that we will obtain, corresponding to the progressive changes in the three interpretation variable sets ($y_2$, $z_1$, and $z_3$), each set of which is  associated with one of the four different representations. 

We have not yet defined the term ``most suitable,'' nor the equivalent term ``best fit,'' also used in this manuscript.

As a starting point, we identified the approximate \textit{h-values} associated with each of interpretation variables for each of these representations. This gave us a potential \textit{h-value} range for each of the four representations. (An illustration of this was provided in Table~\ref{config-var-values-highly-massed-top-rep}, for the data representation of the most agglutinated pattern, shown previously in Figure~\ref{fig:Rocks-three-topographies-horiz-crppd-2020-11-16}\textbf{(a)}.) 

Table~\ref{h-value-range-three-image-reps} gives the low and high ends for the \textit{h-values}, for each of the four representations. These values are approximations, and are obtained via visual inspection when each of the interpretation variables (for each of the distinct representations) has been placed on the analytic graph for the functions $y_2(h)$,  $z_3(h)$, and $z_1(h)$, as shown previously in Figure~\ref{fig:2D-CVM_Expts-scale-free-h-eq-1pt65-rev_2021-02-16_scale-free-rslts}.  

Our task now is to find the \textit{h-value} that provides the ``best fit'' for each of these representations. 

The task of identifying a parameter set yielding the ``best fit'' model for a given data set (or representation of data) is well-known. In fact, the Kullback-Leibler divergence was invented (Kullback and Leibler (1951) \cite{Kullback-Leibler-1951-Info-and-sufficiency}) to address just such a challenge. 

While we can take the Kullback-Leibler (K-L) divergence as an inspiration, or starting point, the K-L divergence is not, in itself, useful for our task. 

Instead, we need a new divergence measure - one specifically devised to deal with the observable variables for a 2-D CVM system.

The following three subsections address this need. First, we review the different notational forms used for the K-L divergence, and then the K-L divergence itself, to establish why it is not suitable for our needs. Then, we suggest a new divergence measure. 

%
\subsection{Notation Used for the Kullback-Leibler Divergence}
\label{subsec:notation-kullback-leibler-divergence}
%

A preliminary (and mostly historical) discussion of notation is not typical in a work such as this. However, there is an odd notation-reversal between the notation used by Friston and colleagues  \cite{Friston_2013_Life-as-we-know-it}, and in prior works by Beal (2014) \cite{Beal_2003_Variational-algorithm-approx-Bayes-inference} as well as Blei et al. (2017)  \cite{Blei-et-al_2017_Variational-Bayes}, versus the notation used by many others. This can pose a trap for the unwary, and indeed, caught this researcher unawares. (That story is reserved for Maren (2022) \cite{AJMaren-Ps-and-QsTech-Notes-2022-001}.) 

In this work, and in related works, we adopt the notation used by Friston and colleagues (op. cit.), as we anticipate that the 2-D CVM can play a useful role in active inference, and refer back to Figure~\ref{fig:Fristons-notation_2021-07-27_crppd}, presented in Subsection~\ref{subsec::background-act-inf}. (Sajid et al. (2020) present a useful discussion of active inference vs. reinforcement learning \cite{Sajid-and-Friston-et-al_2020_Active-Inference}.)

To maintain this correspondence, following work by Friston et al. (op. cit.), we use \textit{Q} to refer to the probability distribution of the data itself, or more specifically, in the \textit{representation} of the external system $\Psi$, that is, $Q(\Psi)$. We use \textit{P} to refer to the model of \textit{Q}. In our case, \textit{Q} is the representation grid, e.g. the various patterns (1-4) identified in the previous Section~\ref{sec:data} on Data. The model \textit{P} is the 2-D CVM system \textit{brought to a free energy minimum}.  

The most important notational elements are as follows: 

\begin{enumerate}
\item The set of model-based observation probabilities is denoted by \textbf{\textit{P}}, or by  \textbf{\textit{p}} for a localized model prediction, and 
\item The set of actual observations is represented by  \textbf{\textit{Q}}, or by \textbf{\textit{q}} for a localized observation.  
\end{enumerate}

It is important to note that \textbf{\textit{this is a reversal}} of how the \textbf{\textit{P}} and \textbf{\textit{Q}} notations are used in many other descriptions of the K-L divergence.  (See examples in Maren (2022) \cite{AJMaren-Ps-and-QsTech-Notes-2022-001}.) 

In the context of active inference, Friston et al. (op. cit.) use a more extended notation. (See Figure~\ref{fig:Fristons-notation_2021-07-27_crppd} for an illustration.) The following points are extracted from Maren (2019a) \cite{ Maren_2019_Deriv-var-Bayes}. 

Key notation definitions are as follows (drawing from the work of Friston et al. (op. cit.): 

\begin{enumerate}
\item \textbf{\textit{The external system $\tilde{\Psi}$}}, which is composed of units $\tilde{\psi}$; we are trying to model this, and construct a \textit{representation} of the external system to do this; in the context of the current work, we have three images of the external system $\tilde{\Psi}$, presented in Figure~\ref{fig:Rocks-three-topographies-horiz-crppd-2020-11-16}, 
\item \textbf{\textit{The internal system $\tilde{R}$}}, which is composed of representational units $\tilde{r}$; where the values $\tilde{r}$ are influenced by sensing processes ($\tilde{s}$) of the external system units $\tilde{\psi}$, and in turn have an active influence $\tilde{a}$ on the external system units $\tilde{\psi}$ (we are temporarily ignoring $\tilde{s}$ and $\tilde{a}$); for this work, our representational units $\tilde{r}$ are expressed as the 2-D CVM grid with active units \textbf{A} and inactive units \textbf{B}, and 
\item \textbf{\textit{$P$ , $Q$, and $\theta$}}: \textbf{\textit{$P$}} is the model of the external system expressed via the internal system, $p$, where the chief distinction is that when we take an actual value for $p$, we do so with the presumption that the internal system is brought to a free energy equilibrium for a given set of parameter values $\theta$. The actual initial data values are represented by $Q$, where individual items are represented as $q$. $\theta$ is the set of model parameters. 
\end{enumerate}

For our purposes (and to make the correspondence with Friston's work clear): 

\begin{enumerate}
\item \textbf{\textit{$Q$}} is the initial data representation; it is the set of configuration variables obtained when we apply a 2-D grid to an image, such as those presented in Figure~\ref{fig:Rocks-three-topographies-horiz-crppd-2020-11-16}; the actual elements of $Q$ are the initial set of configuration variables counted from that grid (including wrap-arounds), are are here denoted as the set $\{q\}$, 
\item \textbf{\textit{$\tilde{P}$}} is the corresponding set of model values, which we further denote as the set $\{p\}$, and these are obtained by identifying the configuration variable values once we bring the initial representation to a free energy minimum, and 
\item \textbf{\textit{$\theta$}} is the parameter set that we are assessing over a certain range, in order to find those parameters that yield the minimal divergence between the model values and the initial representation values; for our purposes, this is the pair of parameters  $(\varepsilon_0, h)$, and for the specific experiments conducted here, we are focused exclusively on $h$, as we have $\varepsilon_0 = 0$.
\end{enumerate}

Note that the ``tilde'' notation was introduced by Friston et al. in his 2015 work \cite{Friston-et-al_2015_Knowing-ones-place-free-energy-pattern-recognition}, but was not in the predecessor 2013 work \cite{Friston_2013_Life-as-we-know-it}. It refers to the notion that all these variables are ``generalized'' variables. The tilde notation can be dropped, with no loss of meaning, in the rest of this work. See Maren (2019a) for further details and interpretation \cite{ Maren_2019_Deriv-var-Bayes}.

%
\subsection{The Kullback-Leibler Divergence}
\label{subsec:kullback-leibler-divergence}
%

One of the most  popular methods -- indeed a bulwark -- for identifying the ``best'' set of model parameters is the Kullback-Leibler divergence \cite{Kullback-Leibler-1951-Info-and-sufficiency}. We begin with the Kullback-Leibler divergence as a starting point, but move (in the following subsection) to a new divergence method; one more suitable for working with comparisons of model versus representation systems for 2-D topographies. 

The following Eqn.~\ref{eqn:variational-K-L-divergence-summation} expresses the Kullback-Leibler divergence using the notation adopted in Friston (2013) \cite{Friston_2013_Life-as-we-know-it}. This same equation is also used in Maren (2019a) \cite{Maren_2019_Deriv-var-Bayes}, which provides a detailed discussion of the variational Bayes method, especially in the context used by Friston.

\begin{equation}
\label{eqn:variational-K-L-divergence-summation}
  \begin{aligned}
   D_{KL}[q({\tilde{{\psi}}}|\tilde{r})||
    p(\tilde{{\psi}}|\tilde{s},\tilde{a},\tilde{r})] = \sum_{i=1}^I q({\tilde{{\psi}}}|\tilde{r}) \ln\left({\frac {q({\tilde{{\psi}}}|\tilde{r})}
    {p(\tilde{{\psi}}|\tilde{s},\tilde{a},\tilde{r})}}\right).    
  \end{aligned}
\end{equation}

The following is extracted from Maren (2019a), Subsection 4.1, ``Interpreting the K-L Divergence.''

``We briefly interpret the physical meaning of the terms in Eqn.~\ref{eqn:variational-K-L-divergence-summation}. The K-L divergence measures the difference between the model (i.e., probability distribution over) of the external system, $p$, and the external system itself, $\tilde{\psi}$.'' (Note: the notation is changed from that in the currently-published version of Maren (2019a); that document will be updated with the switch in $P$ and $Q$ notation as soon as this document is complete.) 

Returning to the extract from Maren (2019a): 

``The previous Eqn.~\ref{eqn:variational-K-L-divergence-summation} includes a summation sign, which is typically found in expressions of the K-L divergence. This summation, however, refers to summing over all instances of data points in the system being modeled (here, denoted $\tilde{\psi}$, as it occurs with a specific probability $q$) and the corresponding points in the model, denoted  $p(\tilde{\psi})$. 

``The model $p$ is a model of the external system, $\tilde{\psi}$, which is why we write $p = p(\tilde{\psi})$. The key feature in computing $p$ is that (for the application being considered here) we take it at the equilibrium state. That is, $p$ corresponds to the \textit{equilibrium free energy} of the external system, which can be computed (or approximated) if we have a suitable free energy equation. Thus, in Eqn.~\ref{eqn:variational-K-L-divergence-summation}, we are looking at the divergence between the model of the system at equilibrium and the probabilities of various components of the system, potentially in a not-yet-at-equilibrium state. 

``The parameter(s) $\theta$ can indeed influence $p$, but the notation for $\theta$ is suppressed throughout. 

``Thus, we can read the term $p(\tilde{\psi}|\tilde{r})$ as the `probability distribution of the model of the external system $\tilde{\psi}$, which is computed based solely on the value of the representational units $\tilde{r}$ that are isolated from the external system $\tilde{\psi}$ by a Markov blanket, but these representational units are to be considered with their at-equilibrium values.''

%
\subsection{Reinterpreting the Kullback-Leibler Divergence}
\label{subsec:reinterp-kullback-leibler-divergence}
%

When we step back and examine Eqn.~\ref{eqn:variational-K-L-divergence-summation}, we see that the summation in Eqn.~\ref{eqn:variational-K-L-divergence-summation} expresses the sum over distinct data quanta. 

For the case that we are addressing here, it is neither sensible nor practical to address individual nodes and individual configuration elements within the 2-D topographies. Instead, it makes more sense to work with the configuration variable values. 

First, we make a simplification in our notation. Instead of the more formally accurate notation involving $\tilde{\psi}$, $\tilde{s}$, $\tilde{a}$, and $\tilde{r}$, we will take it as understood that we are constructing a model of the data representation $\tilde{r}$. Further, we will omit the \textit{tilde} notation going forward. 

If we were concerned only with the distribution of nodes in ``on'' and ``off'' states; i.e., measuring only $x_1$ and $x_2$, then rewriting Eqn.~\ref{eqn:variational-K-L-divergence-summation} would give us

\begin{equation}
\label{eqn:variational-K-L-divergence-summation-simplified}
  \begin{aligned}
   D_{KL}[q(r)|| p(r)] = \sum_{i=1}^2 x_{i,q} \ln\left({\frac {x_{i,q}} {x_{i,p}}}\right).    
  \end{aligned}
\end{equation}

In this case, the summation would be over two states, and we would have $q(r) = x_1, x_2$ in the topography that we are modeling, and $p(r) = x_1, x_2$ in the resultant, free-energy-minimized topography (the model). For clarity, we could identify these as $p(r) = \{x_{1,p}, x_{2,p}\}$ and $q(r) = \{x_{1,q}, x_{2,q}\}$. The associated parameter set is given as $\theta = \{\varepsilon_0,\varepsilon_1\}$.

If we were to apply this to the natural topographies that we have selected for this work, the divergence value that would be found by applying Eqn.~\ref{eqn:variational-K-L-divergence-summation-simplified} would yield a value of zero, regardless of the \textit{h-value} used. This is because by selecting an equiprobable distribution of units, we are constraining that $p(r) = q(r) = 0.5$ for both the ``on'' and `'off'' states. 

Clearly, this Eqn.~\ref{eqn:variational-K-L-divergence-summation-simplified} would be neither sufficient nor appropriate for our needs.

To work with natural topographies and their represenations, or with any 2-D system where the interest is in local topographies, we need to include terms indicative of relations between the remaining configuration variables. 

To do this, we introduce a new divergence measure, expressed as 

\begin{equation}
\label{eqn:variational-K-L-divergence-summation-config-vars}
  \begin{aligned}
   D_{2D-CVM}[q(r)|| p(r)] = \\
 &   2 \sum_{i=1}^3  \beta_i y_{i,q} \ln\left({\frac {y_{i,q}} {y_{i,p}}}\right) + \sum_{i=1}^3  \beta_i w_{i,q} \ln\left({\frac {w_{i,q}} {w_{i,p}}}\right) \\
 &  - \sum_{i=1}^2 x_{i,q} \ln\left({\frac {x_{i,q}} {x_{i,p}}}\right) -  2 \sum_{i=1}^6 \gamma_i  z_{i,q} \ln\left({\frac {z_{i,q}} {z_{i,p}}}\right)   
  \end{aligned}
\end{equation}

We used this measure to obtain a set of divergences between the model (for each of a specific set of \textit{h-values}) and the initial, corresponding configuration variables. 

%
\subsection{Achieving a Free Energy Minimum}
\label{subsec:achieve-free-energy-min}
%

For each \textit{h-value} that we used within a given test range, we used the simple pair-flipping strategy as was described in  (Maren, 2019b, 2021) \cite{Maren_2019_2D-CVM-FE-fundamentals-and-pragmatics, Maren_2021_2-D-CVM-Topographies}. We limited the total number of potential node-state flips to 100 per trial. Typically, only a fraction of the attempts to find a pair of nodes where flipping resulted in a free energy decrease. Even with that, the free energy minimum that could readily be achieved for a given trial was obtained well before the limit of 100 trial ``flips'' was reached.

%
\section{Results}
\label{sec:results}
%

The results, shown in Table~\ref{h-values-for-four-representation-topographies}, are pleasantly in accordance with expectations. 

Specifically, we note the following:

\begin{enumerate} \itemsep0pt 
\item \textbf{\textit{Progression of h-values}} -- the \textit{h-values} progress smoothly in the manner expected, 
\item \textbf{\textit{Configuration (interpretationi) variables progress smoothly as expected}} -- the interpretation variable values ($y_2$, $z_1$, and $z_3$) progress in the manner expected, and
\item \textbf{\textit{Configuration variables diverge from analytic results}} -- while the overall progression of \textit{h-values} is as expected, and the interpretation variable values for each of the free energy-minimized states correspond to approximately the same \textit{h-value}, there is still a surprisingly strong divergence from the analytically-predicted \textit{h-value} anticipated for a given set of interpretation variables and the \textit{h-value} that is obtained computationally; this was noticed in earlier works as well (see Maren (2019b, 2021) \cite{Maren_2019_2D-CVM-FE-fundamentals-and-pragmatics, Maren_2021_2-D-CVM-Topographies}). 
\end{enumerate} 

The deviance of the computationally-obtained \textit{h-value} from the analytically-predicted becomes more pronounced as we move to topographies that are most agglutinated; e.g., as we move from Patterns  \textbf{3a} and  \textbf{3b} (most granular,  (corresponding to Figure~\ref{fig:Rocks-three-topographies-horiz-crppd-2020-11-16}\textbf{(c)}) to Pattern  \textbf{1} (corresponding to Figure~\ref{fig:Rocks-three-topographies-horiz-crppd-2020-11-16}\textbf{(a)}).

As a specific example, for Pattern \textbf{3b}, the interpretation variables are all closely aligned with an approximate \textit{h-value} of $h=1.16$, but the actual \textit{h-value} that gave the smallest (in magnitude) divergence was $h = 1.30$. 

 For Pattern \textbf{1}, which has the largest ``masses'' of \textbf{A} nodes, the \textit{h-value} that provides the smallest divergence is $h = 1.95$. This value, however, is literally ``off the charts'' when we refer to the graph of the analytic solutions for the interpretation variables as functions of \textit{h}. Instead, when we examine the placement of the three interpretation variables on the analytic graph showing those variables as functions of \textit{h}, we see that they center (loosely) around $h \approx 1.43$. (See the MS PPTX$^{TM}$ in the GitHub repository for the graph.)

This third finding - that the actual \textit{h-values} that provide the smallest divergence vis-a-vis the corresponding initial pattern differ substantially from what would be analytically-predicted - is in keeping with results originally obtained in Maren (2019b, 2021) \cite{Maren_2019_2D-CVM-FE-fundamentals-and-pragmatics, Maren_2021_2-D-CVM-Topographies}. 

This difference between the computationally-obtained and the analytically-predicted \textit{h-values} is likely due, at least in part, to the divergence in the analytic equations themselves, due to the quadratic nature of the denominator. 

\begin{table}[!t]
\renewcommand{\arraystretch}{1.3}
\caption{Minimal-Divergence \textit{h-Values} for Four Naturally-Occurring Patterns}
\label{h-values-for-four-representation-topographies}
\centering
\begin{tabular}{|c||c|c|c|c|c|}
\hline
\textbf{Pattern} &  \textit{\textbf{h-Value}} & \textbf{ Diverg} & \textbf{$y_2$} & \textbf{$z_1$} & \textbf{$z_3$}\\
\hline
\textbf{1} & 1.95 & -0.9124  & 0.1602  & 0.2441  & 0.0605 \\
\hline
\textbf{2} &  1.75 & -1.0311 & 0.1738 & 0.2207  & 0.0723 \\
\hline
\textbf{3a} & 1.35 & -1.3089 & 0.2109 & 0.1738  & 0.0957 \\
\hline
\textbf{3b} & 1.30 & -1.3371 & 0.2324 & 0.1426 & 0.107  \\
\hline
\end{tabular}
\end{table}

%
\section{Discussion}
\label{sec:discussion}
%

This work represents a significant milestone in being able to use the Kikuchi-Brush 2-D CVM as a modeling tool. 

Up until now, any efforts to do this would have been hampered because there was no conclusive method to determine a ``best fit'' for any 2-D topography. Now, the method demonstrated here will make it possible - in the very near future - to fill out a 2-D CVM phase space, identifying the configuration variables and free energies (and other thermodynamic variables) associated with a given $(\varepsilon_0, h)$ parameter set. 

By making the 2-D CVM into a useful tool for topography characterization, we can take steps that were not possible up until now. 

For the sake of brevity, the following is couched in very general terms. More extended versions, with substantial references to published literature, will be presented in future works.

%
\subsection{CORTECONs}
\label{subsec:cortecons}
%

Broadening the scope of this discussion, one of the great limitations in artificial intelligence (AI), and specifically in the use of neural networks, has been the restriction of node-connections to being between extrinsically observable nodes (typically, input / output nodes) and the ``latent variables.'' Associated with that is what this author describes as the ``Principle of Sparse Representation.'' This means, that for very good and substantial reasons, when a neural network is designed, the number of latent variable ``hidden nodes'' needs to be kept as small as possible. This ensures that the network's learning yields good generalizations among the latent variables. Other benefits are associated with this sparseness as well. An organization of neural network architectures, together with insights about design principles, was expressed in 1990 by Maren \cite{AJMaren-CHPTR-Hndbk-Nrl-Cmpt-Applctns_Introduction, AJMaren-CHPTR-Hndbk-Nrl-Cmpt-Applctns_Hybrid-Complex-Ntwks}, and further refined in Maren (1991) \cite{AJMaren-1991-logical-topology-w-access}.

This has led to an architectural evolution - carried through from simple Boltzmann machines to the restricted Boltzmann machine to deep learning - in which there can be many layers, but at each layer, the number of latent variable nodes is kept to a minimal number. 

This has further repercussions. One of the most significant is in how neural networks handle temporal behaviors. Currently, combinations of convolutional neural networks and long short-term memory (LSTMs) neural networks are useful in modeling temporal behaviors. However, the design of  LSTMs is, by necessity, somewhat forced and artificial. This has been a ``necessary evil,'' up until now. 

Now, it is possible to design new architectures. These will, in fact, simply be realizations of an architectural evolution first proposed by Maren et al. in 1992 \cite{Maren_1992_Free-energy-as-driving-function, Maren-Schwartz-Seyfried_1992_Config-entropy-stabilizes, Schwartz-Maren_1994_Domains-interacting-neurons}. This is the class of CORTECON \textit{COntent-Retentive TEmporally-CONnected}) networks, which include (at least one) layer in which there is a 2-D CVM grid. It is likely that most CORTECONs will also include the typical sparse latent variable layer(s), which may or may not be in parallel with the 2-D CVM layer(s). 

One potential architectural step with CORTECONs is to allow lateral (potentially Hebbian) communications between nodes within the 2-D CVM grid layer. This may specifically assist recognition of temporal associations if we allow graceful degradation, over time, of a pattern that has been activated by any one set of inputs. Object-oriented methods will need to be invoked to ensure that the most ``long-lived'' node activations are those that are most interior to a group of nodes. Further, we want a given input pattern to activate only a relatively small number of nodes, so that multiple pattern presentations can activate different sets of nodes within the 2-D CVM layer. 

Various communication strategies should be possible between these nodes. There is already a substantial body of work on message-passing, and prior work by Yedidia et al. (2002) \cite{Yedidia-Freeman-Weiss_2002_Understanding-belief-prop}, Wainwright and Jordan (2008) \cite{Wainwright-and-Jordan_2008_Graph-models-exp-fam-var-inf}, and Parr et al. (2019) \cite{Parr-et-al_2019_Neuronal-messaging} discuss message-passing methods that reference the cluster variation method. A limited study by this author (not reported in the literature) suggests that applying belief propagation methods, of any sort, to a 2-D CVM grid system may not be as easy to implement as might be desired. However, this remains a potential avenue for investigation. 

A further discussion of CORTECON principles and architectures is deferred to follow-on works.

%
\subsection{Active Inference}
\label{subsec:active-inf-discussion}
%

Sajid et al. (2020) have stated that ``in active inference an agent’s interaction with the environment is determined by action sequences that minimize expected free energy (and not the expected value of a reward signal).'' ... (p.5) \cite{Sajid-and-Friston-et-al_2020_Active-Inference}. 

What the 2-D CVM offers, that makes it interesting and potentially useful with regard to other models, is that the 2-D CVM is \textit{in itself} brought to a free energy minimum during the course of modeling. Thus we can envision scenarios in which free energy minimization, long advocated by Friston and colleagues as a fundamental process within the brain (op. cit.), is part of obtaining a desired reward. 

Studies by Sajid et al.\cite{Sajid-and-Friston-et-al_2020_Active-Inference}, and previously by Cullen et al. \cite{Cullen-and-Friston-et-al_2020_Active-Inference-original-study} (both associated with Friston's group), have typically used a discrete-state-space system (e.g., the game of Doom). These studies have made it possible for the active agents to epistimologically explore their environments. If a 2-D CVM were to be used, the environments open to active agents could be expressed in a more abstract sense. 

As a further step, it would be possible to guide the agent's traversal through its environment by changing either or both of the parameters in the set $\{\varepsilon_0, h\}$. This would make it possible to specify trajectories through phase space, influencing the agent's actions over time.

%
\section{Summary and Conclusions}
\label{sec:summary}
%

The primary value of this work lies in demonstrating a method by which  a parameter set can be selected so that the resulting 2-D CVM model  provides the most suitable correspondence to a given topography representation. In the demonstration provided here, the parameter set $\{\varepsilon_0, h\}$ is reduced simply to the \textit{h-value}, as $\varepsilon_0$ is set to zero. The demonstration, done on four representations drawn from three different (but related) natural topographies, yields \textit{h-values} whose progression follows that which would be expected, based on the change in the four topography representations. 

In this work, we have introduced a new method for modeling 2-D topographies, using the 2-D CVM (cluster variation method), introduced by Kikuchi in 1953, and where the specific 2-D CVM that we address here was developed by Kikuchi and Brush in 1967 \cite{Kikuchi-Brush_1967_Improv-CVM}. The challenge has been to identify the parameter set yielding the ``best match'' to a given initial topography. 

We created a simple dataset of three different initial 2-D topographies, selected to illustrate different distributions of configuration variables. The premise was that each of these initial topograhies would be at- or close-to equilibrium, as they were naturally-occuring. We created a representation of each of these topographies, with the condition that each resulting representation had to have equiprobable numbers of ``on'' (black lava) and ``off'' (white coral) units. For the last dataset (``Topography \textbf{(c)}''), we were able to extract \textbf{\textit{two}} patterns (Patterns 3a and 3b). This was possible because in this last topography, there was less interactive energy, so the ``clusters'' were much smaller than were observed in Topographies 1 and 2. Thus we were able to shift the representation grid (the grid that we overlaid on the actual image to identify node activations for each of 256 nodes, with the stipulation that there would be equal numbers of nodes in states \textbf{A} and \textbf{B}), to extract two patterns from the same starting image. 

We anticipated that the \textit{h-values} for these last two patterns, drawn from the same originating topography image, would be very close to each other. Specifically, they would be closer to each other than they would be to the \textit{h-values} for Patterns 1 and 2.  

This expectation was met, as evidenced in Table~\ref{h-values-for-four-representation-topographies}. 

The rationale for enforcing equiprobability of ``on'' and ``off'' units in the constructed representations was twofold. First, it allowed comparison of the representational configuration variables with the analytic solutions. Even though previous work by Maren (2019b, 2021) \cite{Maren_2019_2D-CVM-FE-fundamentals-and-pragmatics, Maren_2021_2-D-CVM-Topographies} showed that the analytic solution did not hold far from the case where the interaction enthalpy parameter diverged from zero (\textit{h-value} diverged from 1), there would be at least a starting reference point. 

Second, by enforcing equiprobability of the bistate units, we could address the simple case where the activation enthalpy parameter ($\varepsilon_0$) is zero. That meant that we were seeking only to find the ``best'' \textit{h-value} corresponding to each topography representation, and not the parameter pair \textit{($\varepsilon_0$, h-value)}. This resulted in a simpler analysis, and made it easier to assess the resulting \textit{h-values}. 

The work described here is best viewed as the second in a series, where the first paper (Maren, 2021 \cite{Maren_2021_2-D-CVM-Topographies}) presented initial topographic studies and correlations between \textit{h-values} and both topographies and configuration variables. (Maren (2019b) gave a more extended version of the same \cite{Maren_2019_2D-CVM-FE-fundamentals-and-pragmatics}.)

There were three key findings from that  investigation, which we can update now in light of results obtained from this work:

\begin{itemize}
\setlength{\itemsep}{1pt}
\item \textbf{Analytic vs. computational results} - we continue to see a divergence in the computationally-obtained \textit{h-values} versus those analytically-predicted for a given set of configuration variables,  
\item \textbf{Useful ranges of the $\varepsilon_0$ and $\varepsilon_1$ (\textit{h-value}) enthalpy parameters} -  based on prior work (Maren, 2019b \cite{Maren_2019_2D-CVM-FE-fundamentals-and-pragmatics}), it is still reasonable to keep $\varepsilon_0 < 3$, and possibly smaller; in this work, we found that an \textit{h-value} of $h=1.95$ provided the best model for the most agglutinated pattern used, so an \textit{h-value} in the neighborhood of $h \approx 2$ remains the current high limit for \textit{h}, and
\item \textbf{Useful results are most readily obtained when the initiating pattern is already somewhat close to an equilibrium state} - the prior work used manually-devised patterns, and although each was brought to what appeared to be a free energy-minimized state, the algorithm used to do so was likely not able to bring either pattern to the lowest free energy possible (within a reasonable number of steps); this was evident in that the resulting configuration variable values did not line up on the analytic graph. In the current effort, starting with grids representing near-equilibrium patterns, we saw that the resulting configuration variable values more closely aligned with a single analytic \textit{h-value} for each starting grid. This is encouraging, even if the actual \textit{h-values} diverged from the analytic.  
\end{itemize}

In addition to these basic findings, we continued to see interesting topographic features in the 2-D topographic grids that were generated during the free energy minimization process; these included topographic features such as ``spider-legs,'' ``channels,'' and ``rivers.'' 

We note that it should now be possible to easily populate the $(\varepsilon_0, \varepsilon_1)$ phase space.

%
\section{Future Directions}
\label{sec:future-directions}
%

The predecessor paper identified three areas for \textit{future directions}, and they continue to hold as necessary next-steps: 

\begin{itemize}
\setlength{\itemsep}{1pt}
\item \textbf{Map the phase space} - we need the full set of configuration variable values associated with parameter sets $(\varepsilon_0, \varepsilon_1)$. 
\item \textbf{Smart strategies} - once we know the target configuration variables associated with a given parameter set, we can evolve smart strategies to move in the direction of a known solution. 
\item \textbf{Topography characterization} - as we understand the topographies induced by different parameter sets, we can begin correlating free energy-based topographies with various kinds of observable 2-D patterns, ranging from physical landscapes to medical images. 
\end{itemize}

We briefly discuss each of these.

%
\subsection{Phase Space Mapping }
\label{subsec:phase_space-mapping}
%

The most pressing task now is to map out the $(\varepsilon_0, h)$ (or perhaps more practically, the $(\varepsilon_0, h)$) phase space. The objective is that given a parameter pair  $(\varepsilon_0, h)$, it will be possible to identify the associated set of configuration variables $\{C\}$, and in fact, the free energy for the free-energy-minimized system associated with those configuration variables. 

With the method provided in this work, it will be straightforward to map a set of configuration variables (and associated thermodynamic values) to a given \textit{h-value} (or $\varepsilon_1$), for any point along the $\varepsilon_0 = 0$ axis. 

The real work will be to map the interior phase space, i.e., where $\varepsilon_0 \neq 0$. The code and methods used here for the case where $\varepsilon_0 = 0$  can be readily adapted for this next step. 

It is reasonable to assume that this phase space will be largely continuous in nature, although there may be certain phase space boundaries or discontinuities. We would anticipate that the minimal free energy, at least as a function of $x_1$, will be similar in the 2-D CVM as for the corresponding system in which only the relative fraction of nodes in states \textbf{A} and \textbf{B} is addressed. There are known phase space boundaries (Maren, 1984 \cite{Maren-et-al_1984_Theoretical-model-hysteresis-solid-state-phase-trans}), and it will be interesting to trace the correlation from the simple Ising model to the 2-D CVM model. 

Beyond that, it will be important to investigate the behavior of the free energy \textit{around} points in the phase space. For example, if we have a free energy minimum, it will be important to know things such as: is this a shallow/broad minimum, or is it very well-defined? This will impact how readily the associated parameter range can be easily used in modeling. 

It may be possible to model the phase space using a variational autoencoder (VAE), as suggested by Walker et al. (2020) \cite{Walker-Tam-and-Jarrell_2020_Deep-learning-2-D-Ising-var-crossover}. That would be an entirely different investigation, and should be addressed only when the phase space is substantially mapped out.

%
\subsection{Smart Strategies }
\label{subsec:smart-strategies}
%

As mentioned in the conjoined predecessor works by Maren (2019b, 2021) \cite{Maren_2019_2D-CVM-FE-fundamentals-and-pragmatics, Maren_2021_2-D-CVM-Topographies}, one of the important tasks beyond phase space mapping is to form \textit{smart strategies} for moving from an initial pattern towards a free energy-minimized state. The current strategy is the simplest-possible one of randomly selecting two nodes (of different activations), swapping their activation states, and testing to see if the free energy has been reduced. 

As we develop smart strategies, it will be possible to move through the phase space as adroitly as possible when there are parameter changes. This will further enable a system to follow a phase space trajectory. 

Moving to an object-oriented approach in the 2-D CVM code will be an important element of creating these strategies. It will also enable more complex CORTECON behaviors, such as slow decay around the edges of a large mass, or rule-based Hebbian-like connections between nodes.

%
\subsection{Topography Characterization}
\label{subsec:Topography-characterization}
%

Just as Figure~\ref{fig:Rocks-three-topographies-horiz-crppd-2020-11-16} illustrated a continuum of topographies, a~potentially fruitful line of work will be to address how 2-D CVM networks correspond to previously-identified topography types, whether they are extant in nature or are more abstract.

One potentially useful task in this realm will be to investigate how the 2-D CVM can be used to characterize fractal topographies, or to identify the extent to which scale-invariant configuration variables persist in a given topography \cite{Lin-et-al_2017_Fractal-topography-scale-invariance}.

%
\subsection{Beyond the Phase Space}
\label{subsec:beyond}
%

The directions identified in the previous three subsections are largely focused on making the full 2-D CVM phase space practical and available, and then, characterizing topographies and making it possible to move smoothly throughout both the phase space as well as a given topography. 

These are all predecessor steps, and will set the stage for the more substantial applications envisioned with development of both CORTECONs as well as use of the 2-D CVM for active inference. 

The first of the future directions identified here, that of phase space mapping, can be done rapidly. It can, in fact, be a ``crowd-sourced'' effort, as the necessary code is available, together with four initial grid patterns that can be gently ``tweaked'' to provide starting points for obtaining new sets of $(\varepsilon_0, h)$ correspondences to configuration and thermodynamic variables. 

Themesis intends that the 2-D CVM phase space be publicly available, and welcomes contributions to collaborative efforts.

%
\section*{Code and Data Availability}
\label{sec:code-and-data}
%

%
\subsection*{Codes, Data, and Results Availability}
\label{subsec:codes-data-results-availability}
%

The codes used in this work are all written in Python 3.6, and are are available in the public GitHub repository: \textbf{ajmaren/2-D-CVM-w-Var-Bayes}. The four patterns used for this work are all encapsulated within the codes themselves. Pattern selection can be changed by the user within the program, and the user can also change numerous control variables. 

Directions for use of the associated codes will be put into a separate MS PPTX$^{TM}$ slidedeck and added to the GitHub repository after this work is published. 

\vspace{5mm} 

\textbf{Python Codes:}

\begin{itemize}
\item \textbf{2D-CVM-def-pttrns-anlytc-FE-var-Bayes-v1pt11-2022-08-25.py} - free energy minimization for a pre-defined pattern, across a range of \textit{h-values}, with a summary of divergences at the end of the code; the user can select which of four pre-defined patterns are used (the natural topographies described in this paper, see Data section) along with the range of \textit{h-values}. 

\end{itemize}

\textbf{Data Preparation and Detailed Results:}

\begin{itemize}
\setlength{\itemsep}{1pt}
\item \textbf{2D-CVM-Var-Bayes-Results-2022-08-17.pptx} - MS PPTX$^{TM}$ file documenting experimental trials, with a summary across all four natural topography representations at the end.
\item \textbf{2D-CVM-Var-Bayes-Data-2022-03-31.pptx} - MS PPTX$^{TM}$ file documenting initial data (the three natural terrains presented in Figure 1) together with data pre-processing steps. 

\end{itemize}

%
\subsection*{Code Updates }
\label{subsec:code-updates}
%

Codes referenced here have been made available through Themesis, Inc. Code updates, more patterns and their results (particularly for the $(\varepsilon_0, \varepsilon_1)$ phase space), and other materials (including documentation) will be available over time.  

Those wishing access to new materials should: 

\begin{itemize}
\setlength{\itemsep}{1pt}
\item Go to www.themesis.com, 
\item  Fill out the ``Opt-In'' form on that page, and  
\item Go to the email address that was used in filling out the ``Opt-In'' form to find a confirmation email, and open and click on that email.
\end{itemize}

Persons doing this will then receive emails from Themesis, Inc. In addition to regular email content, there will be specific emails regarding code updates, bug fixes, new data runs, etc. The available codes are supported by extensive documentation, most of which will be placed in the associated GitHub repository. Additional documentation, updates, and new results will be shared via email to those who have completed the Opt-In process. 

The verification and validation for the initial 2-D CVM codes used to produce results presented here is given in Maren (2018) \cite{AJMaren-TR2018-001v2-V-and-V}.

Inquiries should be directed to: themesisinc1@gmail.com.

%
\section*{Funding and Copyright}
\label{sec:funding-and-copyright}
%

The work presented here was fully supported by Themesis Inc. Internal Research and Development, and is copyrighted by Themesis, Inc. The codes are made available under the MIT Open Source Initiative License (see below). The codes, along with supporting materials, are placed into the public Themesis GitHub repository identified in Subsection~\ref{subsec:codes-data-results-availability}.

\textbf{The codes used here are made available for general use following the MIT Open Source Initiative License:}

\textit{Copyright: 2022 Themesis, Inc.}

\textit{Permission is hereby granted, free of charge, to any person obtaining a copy of this software and associated documentation files (the "Software"), to deal in the Software without restriction, including without limitation the rights to use, copy, modify, merge, publish, distribute, sublicense, and/or sell copies of the Software, and to permit persons to whom the Software is furnished to do so, subject to the following conditions:}

\textit{THE SOFTWARE IS PROVIDED "AS IS", WITHOUT WARRANTY OF ANY KIND, EXPRESS OR IMPLIED, INCLUDING BUT NOT LIMITED TO THE WARRANTIES OF MERCHANTABILITY, FITNESS FOR A PARTICULAR PURPOSE AND NONINFRINGEMENT. IN NO EVENT SHALL THE AUTHORS OR COPYRIGHT HOLDERS BE LIABLE FOR ANY CLAIM, DAMAGES OR OTHER LIABILITY, WHETHER IN AN ACTION OF CONTRACT, TORT OR OTHERWISE, ARISING FROM, OUT OF OR IN CONNECTION WITH THE SOFTWARE OR THE USE OR OTHER DEALINGS IN THE SOFTWARE.}

 Further, a  non-exclusive and irrevocable license to distribute this article is granted to \textit{arXiv}.  

 \vspace{10 pt}


%
\appendix
\section{Appendix A: The 2-D CVM Equation}
\label{sec:Appendix-A-CVM-2-D-Eqn}



%

\renewcommand{\theequation}{A-\arabic{equation}}
\setcounter{equation}{0}  

This appendix recapitulates the free energy formalism for the 2-D CVM grid  originally made by Kikuchi in 1951 \cite{Kikuchi_1951_Theory-coop-phenomena}, and further refined by Kikuchi and Brush in 1967 \cite{Kikuchi-Brush_1967_Improv-CVM} (Eqn. I.16). 

The 2-D CVM has an exact solution when $x_1 = x_2 = 0.5$. The results were iniitially presented, without detail, in Kikuchi and Brush [1967] \cite{Kikuchi-Brush_1967_Improv-CVM} (Eqn. I.16). The full and detailed derivation was presented in Appendix A of Maren (2019b) \cite{Maren_2019_2D-CVM-FE-fundamentals-and-pragmatics}).

For the convenience of the reader, we briefly present certain key points from that derivation.

We begin with the free energy equation, previously introduced in Subsection~\ref{subsec:2D-CVM-free-energy} as Eqn.~\ref{eqn:Bar-F-2-D-basic-eqn}, and repeated here for convenience as

\begin{equation}
\label{Bar-F-2-D-basic-eqn-appendix}
  \begin{aligned}
\bar{F}_{2-D} = F_{2-D}/N = \\
  & \varepsilon(-z_1+z_3+z_4-z_6) - \bar{S}_{2-D}\\
+ & \mu (1-\sum\limits_{i=1}^6 \gamma_i  z_i )+4 
\lambda (z_3+z_5-z_2-z_4)
  \end{aligned}
\end{equation}

\noindent
where the entropy, previously expressed as Eqn.~\ref{eqn:Bar-S-2-D-basic-eqn}, is also repeated here as

\begin{equation}
\label{Bar-S-2-D-basic-eqn-appendix}
\bar{S}_{2-D} = 
  2 \sum\limits_{i=1}^3 \beta_i Lf(y_i)
          + \sum\limits_{i=1}^3 \beta_i Lf(w_i)
          - \sum\limits_{i=1}^2 Lf(x_i) 
          - 2 \sum\limits_{i=1}^6 \gamma_i Lf(z_i),           
\end{equation}

\noindent
and where $Lf(v)=vln(v)-v$. The Lagrange multipliers are $\mu$ and $\lambda$, and we have set $k_{\beta}T = 1$. 

As noted previously in Section \ref{subsec:2D-CVM-enthalpy}, we are working with two terms in the enthalpy expression; the  activation enthalpy and the interaction enthalpy. 

We base the interaction enthalpy term in this equation on the expression introduced first by Kikuchi and Brush \cite{Kikuchi-Brush_1967_Improv-CVM} (Eqn. I.16), who express the enthalpy for the 2-D CVM as

\begin{equation}
\label{enthalpy-Kikuch-Brush-appendix}  
\bar{H}_{2-D} = 2 \varepsilon_1(2 y_2 - y_1 - y_3) = 2 \varepsilon_1(-z_1 +z_3 + z_4 - z_6),
\end{equation}

\noindent
using the equivalence relations  

\begin{subequations} \label{sub:y-def-set-appendix}
\begin{gather} 
  y_1 = z_1 + z_2  \label{sub:y1-z1-z2-reln-appendixn} \\
  y_2 = z_2 + z_4 = z_3+z_5 \label{sub:y2-z2-z4-and-z3-z5-reln-appendix}\\
  y_3 = z_5 + z_6.  \label{sub:y3-z5-z6-reln-appendix}
\end{gather}
\end{subequations}

This expresses the notion that the interaction enthalpy is identified as twice the value of each nearest-neighbor interaction ($y_i$). (The multiplier in front of the $y_2$ term is due to the double degeneracy of $y_2$.) 

If the interaction enthalpy parameter ${\varepsilon_1}$ is positive, then we reduce the overall free energy by increasing the relative proportion of nearest-neighbor interactions between those that are like each other ($y_1$ and $y_3$, or \textbf{A} - \textbf{A} and \textbf{B} - \textbf{B}, respectively), and decreasing the relative proportion of interactions between unlike nodes ($y_2$, or \textbf{A} - \textbf{B} interactions). Conversely, if the interaction enthalpy ${\varepsilon_1}$ is negative, we minimize the free energy by increasing the proportion of unlike nearest neighbor pairs (increasing $y_2$). When  ${\varepsilon_1} = 0$, the configuration variables should all be at what would be expected from random distribution. Specifically, for the case where $x_1 = x_2 = 0.5$, we would expect that $2y_2 = 0.5$, and $y_1 = y_3 = 0.25$.

%
\subsection{Equivalence Relations Among the Configuration Variables}
\label{subsec:App-A-Equiv-Rel_Config-Vars}
%

For completeness, we present the entire set of equivalence relations among the configuration variables. This means that we express the sets of $x_i$, $y_i$, and $w_i$ in terms of the $z_i$ (Eqns. I.1 - I.4 \cite{Kikuchi-Brush_1967_Improv-CVM}):

For the $y_i$:

\begin{subequations} \label{sub:y-def-set-appendix}
\begin{gather} 
  y_1 = z_1 + z_2  \label{sub:y1-z1-z2-reln-appendixn} \\
  y_2 = z_2 + z_4 = z_3+z_5 \label{sub:y2-z2-z4-and-z3-z5-reln-appendix}\\
  y_3 = z_5 + z_6.  \label{sub:y3-z5-z6-reln-appendix}
\end{gather}
\end{subequations}

For the $w_i$:

\begin{subequations} \label{sub:w-def-set}
\begin{gather} 
  w_1 = z_1+z_3 \\
  w_2 = z_2+z_5 \\
  w_3 = z_4+z_6
\end{gather}
\end{subequations}

For the $x_i$:

\begin{subequations} \label{sub:x-y-z-relations-def-set}
\begin{gather} 
  x_1 = y_1+y_2=w_1+w_2=z_1+z_2+z_3+z_5 \label{sub:x1-y-z-relations-def}  \\
  x_2 = y_2+y_3=w_2+w_3=z_2+z_4+z_5+z_6  \label{sub:x2-y-z-relations-def} 
\end{gather}
\end{subequations}

The normalization is:
\begin{equation}
  1=x_1+x_2 =\displaystyle\sum\limits_{i=1}^6 \gamma_i z_i.
\end{equation}\\ 

These equivalence relations can be used to shift from expressing the interaction enthalpy in terms of the $z_i$ to an expression using the $y_i$, as discussed in Subsubsection~\ref{subsubsec:2D-CVM-interact-enthalpy-eps1}. 

We note that we can express the interaction enthalpy using the triplet configuration variables $z_i$, instead of the nearest-neighbor pair variables $y_i$. We can do this by drawing on \textit{equivalence relations} between the $y_i$ and $z_i$ variables. Those for $y_2$ are given as

\begin{eqnarray}
\label{eqn:equivalence-relationships-y2-and-z}
  y_2 = z_2+z_4 = z_3+z_5 \\
  2 y_2 = z_2+z_4 + z_3+z_5.   
\end{eqnarray}

Notice that we have two ways of expressing $y_2$ in terms of the $z_i$. Since we want to work with the total $2y_2$, it is easy to express that as the sum of the two different equivalence expressions. This will prove useful when we analytically solve for the free energy minimum, or equilibrium point. 

We also have equivalence relations for $y_1$ and $y_3$ (recapitulating equations given earlier), given as

\begin{equation}
\label{eqn:equivalence-relationships-y2-and-z}
 y_1 = z_1 + z_2
\end{equation}

\noindent
and

\begin{equation}
\label{eqn:equivalence-relationships-y2-and-z}
 y_3 = z_5 + z_6.
\end{equation}

This lets us write 

\begin{equation}
\label{Eqn:z3-analyt1-current-approach}
  \begin{aligned}
\bar{H}_{2-D} =
\varepsilon_1(2y_2 - y_1 - y_3) =
\varepsilon_1(z_4+z_3-z_1-z_6). 
  \end{aligned}
\end{equation}

As a minor note, the enthalpy used in previous related work by Maren \cite{AJMaren-TR2014-003, Maren_2016_CVM-primer-neurosci}, was

\begin{equation}
\label{Eqn:z3-analyt2-previous-approach}
  \begin{aligned}
\bar{H}_{2-D} = H_{2-D}/N =
\varepsilon_1(2y_2) = \varepsilon_1(z_2+z_3+z_4+z_5). 
  \end{aligned}
\end{equation}

The results given here are similar in form to the results presented in the two previous works by Maren; they differ in the scaling of the interaction enthalpy term.

%
%


\end{document}